\documentclass[conference]{IEEEtran}
\usepackage{times}

\usepackage{graphics}
\usepackage{graphicx}
\usepackage{pdfpages}
\usepackage{color}
\usepackage{amsmath,amssymb}
\usepackage{mathtools}
\usepackage{color, soul}
\usepackage{multirow}
\usepackage{caption}
\usepackage{multicol}
\usepackage{xcolor}
\usepackage{afterpage}
\usepackage{graphicx}
\usepackage{placeins}
\newtheorem{problem}{\textbf{Problem}}
\newtheorem{assumption}{\textbf{Assumptions}}
\newtheorem{challenges}{\textbf{Challenges}}
\usepackage{dblfloatfix}

\usepackage[font=small,labelfont=bf,margin=\parindent,tableposition=top]{caption}

\usepackage{fancyhdr} 
\usepackage{subcaption}
\usepackage{titlesec}
\usepackage[colorlinks=true, allcolors=black]{hyperref}
\usepackage{dblfloatfix}
\usepackage{float}
\usepackage{chngcntr}
\usepackage{cite}

\titlespacing*{\section}{0pt}{0.5\baselineskip}{0.2\baselineskip}
\titlespacing*{\subsection}{0pt}{0.35\baselineskip}{0.35\baselineskip}

\usepackage[ruled, vlined, linesnumbered]{algorithm2e}

\usepackage{cite}
\usepackage[font=small,labelfont=bf,margin=\parindent,tableposition=top]{caption}
\makeatletter
\let\NAT@parse\undefined
\makeatother

\usepackage[numbers,sort&compress]{natbib}
\usepackage{multicol}
\usepackage{hyperref}
\hypersetup{
    colorlinks=true,
    linkcolor=blue,
    citecolor=blue,
    urlcolor=blue
}

\IEEEoverridecommandlockouts 

\begin{document}

\title{How to Coordinate UAVs and UGVs for Efficient Mission Planning? Optimizing Energy-Constrained Cooperative Routing with a DRL Framework}


\author{Md Safwan Mondal$^{1,\dagger}$, Subramanian Ramasamy$^{1}$, Luca Russo$^{1}$, James D. Humann$^{2}$, \\James M. Dotterweich$^{3}$,  Pranav Bhounsule$^{1}$

\thanks{$^{1}$Md Safwan Mondal, Subramanian Ramasamy, Ragib Rownak, Luca Russo and 
Pranav A. Bhounsule are with the Department of Mechanical
and Industrial Engineering, University of Illinois Chicago, IL,
60607 USA. {\tt\small mmonda4@uic.edu}, {\tt\small sramas21@uic.edu}, {\tt\small lrusso5@uic.edu }, {\tt\small pranav@uic.edu} $^2$James D. Humann is with DEVCOM Army Research Laboratory, Los Angeles, CA, 90094 USA.{\tt\small james.d.humann.civ@army.mil}
   $^3$ James M. Dotterweich, is with DEVCOM Army Research Laboratory, Aberdeen Proving Grounds, Aberdeen, MD 21005 USA. {\tt\small james.m.dotterweich.civ@army.mil}}%
   \thanks{ $\dagger$ Corresponding author, *This work was supported by ARO contract number W911NF-24-2-0018. }
}

\maketitle

\begin{abstract}
Efficient mission planning for cooperative systems involving Unmanned Aerial Vehicles (UAVs) and Unmanned Ground Vehicles (UGVs) requires addressing energy constraints, scalability, and coordination challenges between agents. UAVs excel in rapidly covering large areas but are constrained by limited battery life, while UGVs, with their extended operational range and capability to serve as mobile recharging stations, are hindered by slower speeds. This heterogeneity makes coordination between UAVs and UGVs critical for achieving optimal mission outcomes. In this work, we propose a scalable deep reinforcement learning (DRL) framework to address the energy-constrained cooperative routing problem for multi-agent UAV-UGV teams, aiming to visit a set of task points in minimal time with UAVs relying on UGVs for recharging during the mission. The framework incorporates sortie-wise agent switching to efficiently manage multiple agents, by allocating task points and coordinating actions. Using an encoder-decoder transformer architecture, it optimizes routes and recharging rendezvous for the UAV-UGV team in the task scenario. Extensive computational experiments demonstrate the framework's superior performance over heuristic methods and a DRL baseline, delivering significant improvements in solution quality and runtime efficiency across diverse scenarios. Generalization studies validate its robustness, while dynamic scenario highlights its adaptability to real-time changes with a case study. This work advances UAV-UGV cooperative routing by providing a scalable, efficient, and robust solution for multi-agent mission planning. More details are available on the website: \href{https://sites.google.com/view/muavugvdrl}{ https://sites.google.com/view/muavugvdrl}.
\end{abstract}

\IEEEpeerreviewmaketitle

\section{Introduction}

In applications like disaster response, search-and-rescue operations, and environmental monitoring, rapid and persistent coverage of critical areas is essential. Unmanned Aerial Vehicles (UAVs) play a pivotal role in these tasks, providing real-time surveillance of disaster zones, wildfire perimeters, and critical infrastructure \cite{noguchi2019persistent, lyu2023unmanned, mondal2024robust}. Their ability to access otherwise unreachable areas, such as dense forests, mountainous terrains, and urban rooftops, makes them invaluable for time-sensitive tasks. However, UAVs are fundamentally limited by their battery capacity, requiring frequent recharges thereby disrupting sustained mission coverage \cite{maini2019cooperative}.

To mitigate this limitation, Unmanned Ground Vehicles (UGVs) have emerged as complementary agents in multi-agent systems. UGVs, capable of traversing road networks and operating for extended durations, act as mobile recharging stations for UAVs, extending their operational range and enabling uninterrupted mission execution. The synergy between UAVs and UGVs combining the agility of aerial operations with the endurance and logistical support of ground vehicles has shown immense potential for improving mission efficiency across diverse applications \cite{ding2021review}. However, this demands intelligent coordination to optimize routes, synchronize recharging schedules, and ensure all task points are visited in an efficient manner.

\setcounter{figure}{0} 
\begin{figure}[t]
\centering
\includegraphics[scale=0.4
]{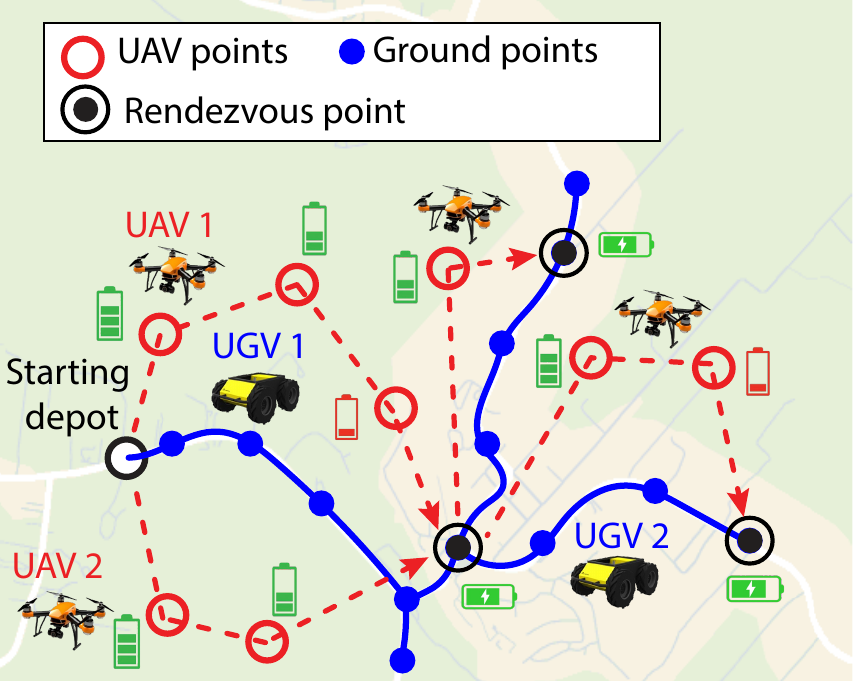}
\caption{Illustration of the fuel-constrained UAV-UGV cooperative routing problem. UAV-specific task points (red hollow circles) and ground task points (blue solid circles) must be visited, with rendezvous points (black circles) for UAVs recharging on the UGVs. The goal is to plan UAV and UGV routes (red dashed and blue solid paths, respectively) to minimize total mission time while adhering to UAV fuel and UGV speed constraints.}
\label{problem}
\end{figure}

The coordination of UAVs and UGVs in such cooperative missions presents unique challenges. UAVs must balance fuel consumption and mission routing (i.e., the order of task point visits) while timing their rendezvous with UGVs for recharging. UGVs, meanwhile, are constrained by their slower speeds and reliance on road networks. In multi-agent systems, these challenges are amplified as multiple UAVs and UGVs must collaboratively plan routes and synchronize operations for seamless task execution. Traditional heuristic-based or hierarchical optimization approaches often struggle to scale with the increasing complexity and dynamic nature of these problems. This highlights the need for a comprehensive cooperative routing framework that optimally plans routes for UAVs and UGVs while synchronizing recharging rendezvous. Such a framework must account for the heterogeneous capabilities of these agents, balance their operational constraints, and deliver robust, scalable solutions that adapt to real-time changes, ensuring efficient and timely mission completion.

\subsection{Related works}

The cooperative routing problem has been extensively explored within the fields of transportation \cite{murray2015flying} and robotics \cite{li2016hybrid, manyam2019cooperative}, with various strategies proposed to address its inherent complexity. These approaches can be categorized into traditional optimization techniques, heuristic methods, and learning-based frameworks, including reinforcement learning for cooperative multi-agent systems. Traditional methods often model UAV-UGV cooperative routing as extensions of the Vehicle Routing Problem (VRP) or the Traveling Salesman Problem (TSP). In transportation, the Truck-Drone Delivery Problem, a specialized form of VRP with Drones (VRP-D) or TSP with Drones (TSP-D) has been formulated using Mixed Integer Linear Programming (MILP) to derive optimal solutions \cite{di2017last, boysen2018drone}. MILP provides a rigorous mathematical framework for integrating energy constraints, time windows, and heterogeneous vehicle characteristics. However, MILP's applicability to large-scale, real-time problems is severely limited due to its computational complexity, as the number of nodes increases \cite{cattaruzza2016vehicle}. Consequently, while MILP offers theoretical optimality, it becomes impractical for dynamic and complex scenarios.

To address these scalability challenges, heuristic and metaheuristic methods have been widely adopted as practical alternatives. Techniques such as Genetic Algorithms (GA) \cite{peng2019hybrid}, Tabu Search (TS) \cite{schermer2019hybrid}, and Simulated Annealing (SA) \cite{liperda2020simulated} have shown effectiveness in approximating optimal solutions within reasonable computation times. Adaptive approaches, including Iterated Local Search \cite{ibaraki2008iterated} and Adaptive Large Neighborhood Search \cite{tacs2014time}, further enhance solution quality by iteratively exploring and refining the solution space, balancing exploitation and exploration. In the context of UAV-UGV systems, heuristic methods often employ multi-echelon or multi-level optimization frameworks to simplify coordination complexities. For instance, strategies like “UGV first, UAV second” decompose the problem into manageable subproblems, where rendezvous points are identified using Minimum Set Cover (MSC) formulations and routes are optimized through MILP or metaheuristic algorithms \cite{maini2019cooperative, maini2015cooperation, ropero2019terra}. Seyedi et al. \cite{seyedi2019persistent, lin2022robust} introduced a heuristic for cyclical patrolling by uniform UAV-UGV teams, leveraging spatial partitioning to optimize area coverage. While effective for continuous surveillance, this method may not be well-suited for scenarios involving discrete task points. Despite their practicality, heuristic methods often require significant domain-specific customization, limiting their adaptability to diverse problem settings. Furthermore, their performance in dynamic, real-time scenarios, particularly in large-scale multi-agent systems remains constrained, underscoring the need for more robust and scalable solutions.

In recent years, learning-based approaches have emerged as promising alternatives for solving complex routing tasks. Deep Reinforcement Learning (DRL) has shown the capability to learn optimal policies through environment interaction, eliminating reliance on handcrafted heuristics \cite{fan2022deep, vinyals2015pointer, mondal2024attention, mondalrisk}. Vinyals et al. \cite{vinyals2015pointer} introduced the Pointer Network, which leveraged attention mechanisms to solve the Traveling Salesman Problem (TSP). This framework was later extended by Bello et al. \cite{bello2016neural}, who incorporated reinforcement learning to tackle more complex VRP and TSP variants. Kool et al. \cite{kool2018attention} advanced the field further by proposing an encoder-decoder transformer architecture with multi-head attention layers, achieving state-of-the-art performance on VRP tasks while surpassing traditional heuristics. Zhang et al. \cite{zhang2023coordinated} and Fuertes et al. \cite{fuertes2023solving} developed DRL-based frameworks using transformer networks for multi-agent cooperative routing, demonstrating scalability and efficiency. However, most existing DRL frameworks focus on single-agent or homogeneous multi-agent systems, often overlooking scenarios with heterogeneous agents, such as UAV-UGV systems, which perform distinct roles. Wu et al. \cite{wu2021reinforcement} applied DRL to optimize truck-drone delivery tasks but restricted their model to one-truck-one-drone systems. Fan et al. \cite{fan2022deep} proposed a DRL-based approach for energy-constrained UAVs, assuming fixed recharging locations, which limits flexibility. Furthermore, Ramasamy et al. \cite{ramasamy2022heterogenous, mondal2023optimizing} emphasized the importance of synchronizing recharging instances in cooperative routing, highlighting their impact on mission efficiency.

This paper addresses the existing gaps in UAV-UGV cooperative routing by proposing a DRL framework that leverages an encoder-decoder transformer architecture with policy gradient method. The framework is designed to optimize coordinated routing for energy-constrained UAVs and UGVs, with UGVs acting as mobile recharging stations. It aims to minimize the total mission time while ensuring efficient synchronization between UAVs and UGVs. The primary contributions of this work are as follows: \vspace{1mm} 
\vspace{1mm} \\
\textbf{1.} We formulate the energy-constrained multi-UAV-UGV cooperative routing problem as a bi-level optimization problem for approximate heuristic solutions and as a Markov Decision Process (MDP) to solve it through DRL using an encoder-decoder-based transformer architecture with attention layers.  \\
\textbf{2.} We introduce a one-agent-per-sortie agent selection strategy, enabling the decentralized framework to scale across varying team sizes and configurations. The approach is evaluated on diverse problem sizes and distributions, \textit{showcasing its generalization capability}.\\
\textbf{3.} We compare our method against heuristic-based bi-level optimization strategies for cooperative routing and a learning-based baseline with an alternative agent selection strategy. Our framework demonstrates \textit{superior solution quality and runtime efficiency, emphasizing its effectiveness.}. \\
\textbf{4.} Through a case study, we demonstrate the utility of our framework for online route planning, \textit{adapting to dynamically appearing task points and changing team configurations during the mission}.

\section{Problem Formulation}

\subsection{Problem Overview}

In scenarios requiring rapid assessment, such as disaster relief, infrastructure monitoring, or environmental surveys, points of interest are often dispersed across a region. These points may include locations accessible via road networks (e.g., near roads) or remote and inaccessible areas (e.g., forests, mountainous terrain, or rooftops). To address this challenge, a heterogeneous team of agents with UAVs and UGVs can be deployed. UAVs, with their high mobility and speed, can efficiently access both accessible and remote locations but are constrained by limited battery capacity. Conversely, UGVs operate exclusively on road networks at slower speeds, serving as mobile charging platforms to extend the operational range of UAVs (see Fig. \ref{problem}). By effectively coordinating the operations of UAVs and UGVs, it is possible to maximize mission efficiency and ensure complete coverage of all task points.

The problem can be defined as follows: Given a set of task points \( \mathcal{M} = \{m_0, m_1, \dots, m_n\} \), partitioned into two subsets: ground points \( \mathcal{M}_g \), accessible by both UGVs and UAVs, and aerial points \( \mathcal{M}_a \), accessible only by UAVs and a heterogeneous system comprising UAVs \( \mathcal{A} = \{ u^a_j \,|\, j = 1, \dots, M \} \) and UGVs \( \mathcal{G} = \{ u^g_i \,|\, i = 1, \dots, N \} \), the goal is to determine optimal routes for all agents to visit every task point in the shortest possible time by any of the UAVs. UAVs, with higher speeds (\( v_a \)) but limited fuel capacity $F_a$, periodically rendezvous with UGVs at ground points for recharging. UGVs, operating at slower speeds (\( v_g \)) along road networks, provide logistical support necessary for UAVs to sustain prolonged operations.

\begin{assumption}
The problem is modeled under the following assumptions:
\begin{itemize}
    \item All UAVs and UGVs initiate the mission from the same starting depot.
    \item UAVs require a constant recharging time \( T_R \) during each rendezvous with a UGV.
    \item The mission starts when any of the UAVs or UGVs departs from the depot and concludes when all task points have been visited, with UAVs completing their final recharge on any UGV.
    \item A UGV can provide recharging services to only one UAV at a time.
\end{itemize}
\end{assumption}

\begin{problem}
The objective is to develop a coordinated strategy for UAVs and UGVs to minimize the total mission time. UAVs must operate within energy constraints and periodically recharge by rendezvousing with UGVs, which adhere to speed constraints. The strategy must ensure that all task points are visited in minimum time with the coordination between UAVs and UGVs.
\end{problem}

\begin{challenges}
The problem presents several challenges due to the heterogeneity of the agents and the operational constraints:
\begin{itemize}
    \item Designing optimal routes for UAVs and UGVs to minimize overall mission time while addressing their operational energy and mobility constraints.
    \item Synchronizing UAV-UGV recharging events to avoid delays while balancing safety (failure because of fuel depletion) and efficiency. Frequent recharging ensures safety but prolongs mission time, whereas fewer recharges reduce time but risk fuel depletion and mission failure.
    \item Managing the interdependence of UAV and UGV routes to ensure seamless coordination, particularly in scheduling rendezvous points and allocating recharging tasks.
\end{itemize}
\end{challenges}

\subsection{MDP Formulation}

This cooperative routing problem is modeled as a sequential decision-making process, where agents iteratively select task points to visit or recharge. The problem is formalized as a Markov Decision Process (MDP), characterized by the tuple \( \langle \mathcal{S}, \mathcal{A}, \mathcal{T}, \mathcal{R} \rangle \), with components defined as follows:

1) The \textbf{state space} \(\mathcal{S}\) represents the environment's status at any decision step. At time \(t\), the state \(s_t \in \mathcal{S}\) is defined as \(s_t = (p_t, f_t, q_t)\), where \(p_t = \{x_t, y_t\}\) indicates the current position of the active agent (i.e., the agent taking action), \(f_t\) denotes its remaining fuel level, and \(q_t = \{x^i, y^i, d^i_t\}\) encodes the coordinates and visitation status of all task points \(m_i \in \mathcal{M}\). Here, \(d^i_t = 1\) if task point \(m_i\) has been visited, and \(d^i_t = 0\) otherwise.

2) The \textbf{action space} \(\mathcal{A}\) includes all feasible task points that the active agent can select at each decision step. Actions \(a_t \in \mathcal{A}\) involve either visiting a task point or recharging at a ground point. Ground points \(\mathcal{M}_g\) serve dual purposes: visiting and recharging. The action space is thus:
$
\mathcal{A} = \{\mathcal{M}_g \,(\text{recharging}) \cup \mathcal{M}_g \,(\text{visiting}) \cup \mathcal{M}_a \,(\text{visiting})\}.
$
Infeasible actions are masked based on the current state \(s_t\), ensuring only valid options are available.

3) The \textbf{reward function} \(r_t = r(s_t, a_t)\) is defined as the travel time \(t_{ij}\) between two task points \(i\) and \(j\), plus a fixed recharging time \(T_R\):
$
r_t = t_{ij} + T_R.
$
Here, \(T_R = 0\) for visiting actions. This per-step reward accumulates to form the overall mission cost. The \textbf{return} \(\mathcal{R}\) measures the total mission cost and is defined as the maximum cumulative time taken by any agent and a penalty \(\mathbb{P} = 800\) is added in case of mission failure, yielding the final return:
$
\mathcal{R} = \underset{a \in \mathcal{A} \cup \mathcal{G}}{\max} \sum_{t=0}^{T} r_t^{(a)} + \mathbb{P}.
$

4) The \textbf{transition function} \(\mathcal{T}\) governs how the environment evolves from \(s_t\) to \(s_{t+1} = (p_{t+1}, f_{t+1}, q_{t+1})\) based on action \(a_t\). The agent's position is updated to the selected node: \(p_{t+1} = \{x_{t+1}, y_{t+1}\} \equiv a_t\). The fuel level is updated as \(f_{t+1} = f_t - f_{ij}\) for visiting actions or reset to \(F_a\) for recharging. The visitation status updates as \(d^i_{t+1} = 1\) for the selected point \(m_i\). Agent switching is governed by a sortie-wise selection strategy (see \hyperref[agent_sel_strategy]{Section~\ref{sec4}}). All transitions are deterministic.

\textbf{Objective}:

The goal of the DRL agent is to learn a policy \(\pi_\theta\) that minimizes the expected return \(\mathcal{R}\), corresponding to the worst-case mission time across all agents. Formally:
\[
\pi^* = \arg\min_{\pi} \mathbb{E}_{\pi} \left[ \mathcal{R} \right]
= \arg\min_{\pi} \mathbb{E}_{\pi} \left[ \underset{a \in \mathcal{A} \cup \mathcal{G}}{\max} \sum_{t=0}^{T} r_t^{(a)} + \mathbb{P} \right]
\]
This formulation aligns with the cooperative routing objective: to minimize total mission completion time.

\section{Heuristics Framework}
\label{sec3}

In this section, we formulate the multi-agent cooperative routing problem using a bilevel optimization framework, which serves as a benchmark that can be addressed using heuristic methods with standard solvers.

\subsection{Bilevel Optimization Framework}

The bilevel optimization framework, also referred to as a multi-echelon strategy, provides a hierarchical approach for solving heterogeneous multi-agent cooperative routing problems. This methodology decomposes the problem into two interdependent levels, often guided by sequencing strategies such as `UGV first, UAV second,' `UAV first, UGV second,' or other prioritization routines \cite{maini2015cooperation, ropero2019terra, maini2019cooperative}. At the first (outer) level, the route for one type of vehicle is planned while considering predefined mission parameters and environmental constraints. This planned route serves as a reference or constraint for the second (inner) level, where the routing of the other vehicle type is optimized. The second level incorporates the interactions between the vehicles, addressing critical factors such as task allocation, temporal coordination, and spatial synchronization. By dividing the problem into manageable subproblems, this hierarchical approach simplifies the complexity of multi-agent coordination and enables the application of heuristic or optimization methods tailored to the unique operational characteristics and constraints of each vehicle type.

For the multi-UAV-UGV cooperative routing problem, we adopt the `UGV first, UAV second' strategy \cite{gao2020commanding}. In the outer level, the routes for UGVs are planned, taking into account their slower speeds, restriction to road networks, and role as mobile recharging stations. In the inner level, UAV sorties are optimized based on the UGVs’ preplanned routes, ensuring efficient energy management and synchronized recharging schedules. Both the stages are discussed in detail:


 \subsubsection{Outer level: Determining UGVs' route sorties }

 The UGVs' operational routes within the road network are determined through a two-step hierarchical process. Initially in the first phase, a \textbf{Minimum Set Cover (MSC)} problem \cite{mustafa2010improved} is solved to identify the essential refueling stops \(\mathcal{R}_e\) required to ensure that all task nodes \(\mathcal{M}\) are accessible to UAVs from the road network following the work of \cite{maini2015cooperation}. The task nodes represent target locations, while the UAVs' fuel capacity \(F_a\) defines the coverage radius. The MSC solution provides the minimum set of refueling stops that guarantee UAV coverage for all task nodes. Once the refueling stops \(\mathcal{R}_e\) are identified, the UGV routes are optimized by solving a Traveling Salesman Problem (TSP). This step determines the most efficient sequence for the UGV to traverse the refueling stops, starting and ending at a fixed depot. The TSP optimization minimizes the UGV's total travel distance and provides an ordered route for visiting the refueling stops. In the second phase of the outer level, the problem is divided into \(n\) independent subproblems, where \(n\) corresponds to the number of refueling stops (including the depot). The UGV is assumed to travel only between two consecutive refueling stops within each subproblem. Each task point is assigned to its nearest refueling stop (including the depot) within the coverage radius. In each subproblem, the starting refuel stop acts as the source node, while the next refuel stop serves as the destination node. UAV task points associated with the destination refuel stop are allocated to the corresponding subproblem. This division ensures that each subproblem contains distinct task points and well-defined UGV traversal paths. Further details about the MSC formulation and subproblem division are provided in the \hyperref[appendix]{Appendix}.

 \subsubsection{Inner level: Determining UAVs' route sorties}
 
Following the division of the problem into subproblems and the allocation of tasks, each subproblem is modeled as an Energy-Constrained Vehicle Routing Problem with Time Windows (E-VRPTW). In this formulation, the source and destination refueling stops establish time windows to synchronize the operations of UAVs and UGVs. The time window for the destination refueling stop is calculated based on the UGV’s travel time to that stop. As UAVs can only land on UGVs after their arrival at the refuel stop, the lower bound of the time window is given by:
\[
t_r = \left[ \frac{\| x_r - x_0 \|}{v_g} \right], \quad \forall \ (x_0, x_r) \in \mathbb{S}
\]
where \(x_r\) denotes the destination refueling stop coordinates, \(x_0\) represents the source refueling stop in subproblem \(\mathbb{S}\), and \(v_g\) is the UGV’s speed. If the UGVs arrive at the refuel stops earlier than the UAVs, they are assumed to wait for the UAVs to land for recharging. To account for multiple recharging events, the destination refuel stop and its time window are duplicated, forming a set of refuel stops \( X_r \) where each duplicated instance corresponds to a separate recharging event (\( x_r \in X_r \)).  

The E-VRPTW formulation represents the problem as a graph-based model where task nodes \( \mathcal{M}_s \) within each subproblem, along with the set of refuel stops \( X_r \), form the vertices of the graph: $V = \mathcal{M}_s \cup X_r.$ Edges between nodes \( i \) and \( j \) are defined as: $E = \{(i, j) \mid i, j \in V, i \neq j\},$ with a non-negative arc cost \( t_{ij} \) representing the travel time between nodes. Decision variables \( x_{ijk} \) indicate whether UAV \( k \) travels from vertex \( i \) to vertex \( j \). UAVs begin their routes at the source depot \( S \), visit task points, and recharge at the designated refuel stop \( x_r \in X_r \). The objective function, as defined in Eq. \ref{eq:1}, minimizes the total travel time for all UAVs. Eq. \ref{eq:oe} ensures that each task point is visited exactly once by any UAV. The fuel constraints are enforced through Eq. \ref{eq:3}, which mandates that UAVs are fully recharged at the destination refuel stop, and Eq. \ref{eq:4},  ensures that UAV fuel levels remain within operational limits throughout the mission. Eq. \ref{eq:5} governs fuel consumption and prevents subtour formation by ensuring that the fuel level decreases according to the travel distance. Time constraints are introduced through Eq. \ref{eq:6}, which ensures that UAVs respect the time windows for refueling stops, and Eq. \ref{eq:7}, which updates the UAVs' arrival time at each node based on travel time. The flow conservation constraint in Eq. \ref{eq:8} guarantees route continuity by enforcing that every UAV entering a node must also exit from it. The starting and ending conditions of UAV routes are stated in Eqs. \ref{eq:9}-\ref{eq:10}, ensuring that UAVs initiate their journeys from the depot and conclude them at a refueling stop. Finally, Eq. \ref{eq:11} enforces feasibility conditions, preventing UAVs from selecting actions that would result in infeasible fuel consumption. These constraints collectively ensure that the UAVs operate efficiently while adhering to energy limitations, mission constraints, and coordination requirements within the multi-UAV cooperative routing framework.


\text{Objective: }
\begin{equation}
\min \sum_{k \in \mathcal{A}} \sum_{i \in V} \sum_{j \in V} t_{ij} x_{ijk}  \label{eq:1}
\end{equation}
\text{Constraints:} 
\begin{equation}
\sum_{k \in \mathcal{A}} \sum_{j \in V} x_{ijk} = 1, \quad \forall i \in M_s
 \label{eq:oe}
\end{equation}

\begin{equation}
f^a_{i,k} = F_a, \quad \forall i \in X_r, \quad k \in \mathcal{A} \label{eq:3}
\end{equation}

\begin{equation}
0 \leq f^a_{i,k} \leq F_a, \quad \forall i \in V \setminus \{S, X_r\}, \quad k \in \mathcal{A} \label{eq:4}
\end{equation}

\begin{equation}
\begin{split}
f^a_{j,k} \leq f^a_{i,k} - (f^a_{ij} x_{ijk})  
+ L_1 (1 - x_{ijk}),  \\ \forall i, j \in V \setminus \{S, X_r\}, \ k \in \mathcal{A}
\end{split}
\label{eq:5}
\end{equation}

\begin{equation}
t_{i,k} \geq t_{r,i}, \quad \forall i \in X_r, \quad k \in \mathcal{A} \label{eq:6}
\end{equation}

\begin{equation}
\begin{split}
t_{j,k} \geq t_{i,k} + t_{ij} x_{ijk} - L_2 (1 - x_{ijk}), \ \forall i, j \in V,  \ k \in \mathcal{A} \label{eq:7}
\end{split}
\end{equation}

\begin{equation}
\sum_{j \in V} x_{ijk} = \sum_{j \in V} x_{jik}, \quad \forall i \in V, \quad k \in \mathcal{A} \label{eq:8}
\end{equation}

\begin{equation}
\sum_{j \in V} x_{S,j,k} = 1, \quad \forall k \in \mathcal{A} \label{eq:9}
\end{equation}

\begin{equation}
\sum_{i \in V} x_{i,X_r,k} = 1, \quad \forall k \in \mathcal{A} \label{eq:10}
\end{equation}

\begin{equation}
x_{ijk} = 0, \quad \text{if } f_{ij} > f_{i,k}, \quad \forall i, j \in V, \quad k \in \mathcal{A} \label{eq:11}
\end{equation}


The UAVs' route sorties are derived by solving the E-VRPTW formulation using standard optimization solvers. These solutions provide the sortie end times, denoted as \(t_{\mathcal{R}}\), along with the designated UGVs facilitating the recharging operations. The rendezvous times are subsequently used to calculate both the UGVs' waiting times at refueling stops and the UAVs' waiting times for recharging, as each UGV can accommodate only one UAV at a time during the recharging process. At each rendezvous point, the UAV spends a fixed recharging time \(T_R\) to complete recharging before resuming its assigned mission. After determining the UAV and UGV route sorties for a specific subproblem, the process advances to the next subproblem. This iterative approach continues sequentially until all subproblems have been addressed, ensuring that every task point is visited and the mission objectives are successfully accomplished. A sample solution of UAVs' route sorties is shown in the \hyperref[appendix]{Appendix}.

To establish a non-learning baseline for our cooperative routing problem, we adopt the bilevel optimization framework detailed earlier. This approach leverages the Google OR-Tools\texttrademark \ CP-SAT solver \cite{ORtools}, which employs constraint programming (CP) techniques to solve the E-VRPTW. To enhance the solver's capability and reduce the likelihood of convergence to local optima, the solver can incorporate different metaheuristic methods: 1) \textbf{Guided Local Search (GLS)}, 2) \textbf{Tabu Search (TS)}, and 3) \textbf{Simulated Annealing (SA)}. These metaheuristics improve the solver’s ability to explore the solution space, producing high-quality approximations within significantly shorter runtimes compared to traditional Mixed Integer Linear Programming (MILP) approaches. This baseline serves as a robust benchmark against which the performance of our proposed DRL-based framework is evaluated.

\setcounter{figure}{2} 
\begin{figure*}[]
\centering
\includegraphics[scale=0.55
]{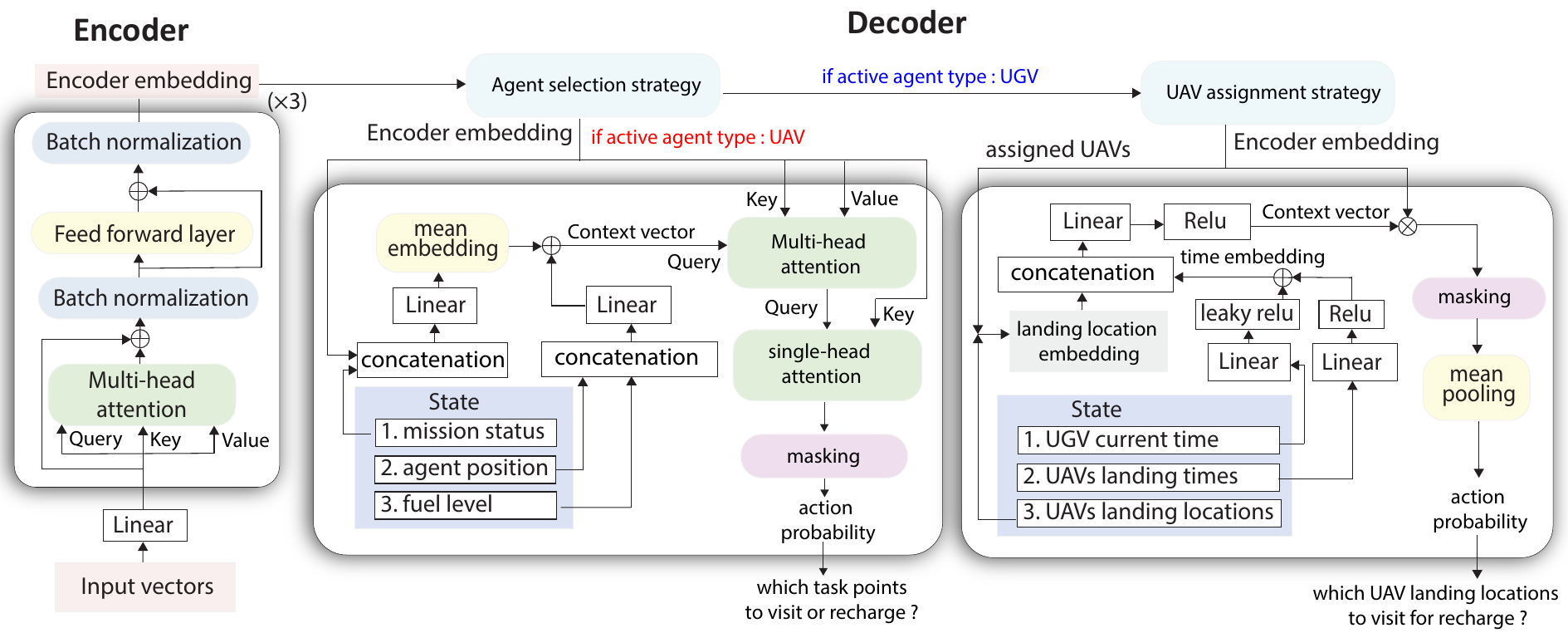}
\caption{Architecture of the proposed transformer network. The encoder consists of three attention layers that process raw input data to generate node embeddings. The decoder employs an agent selection strategy to determine the active agent type (UAV or UGV) and constructs a context vector based on the current state of the selected agent. It uses the input embeddings and the context vector to determine action for the selected active agent.}
\label{architecture}
\end{figure*}

\section{Reinforcement Learning Framework}
\label{sec4}

This section introduces a Deep Reinforcement Learning (DRL) methodology for cooperative routing of multiple UAVs and UGVs. We propose an encoder-decoder transformer-based architecture combined with a reinforcement learning algorithm to learn an optimized routing policy \(\pi_{\theta}\), where \(\theta\) represents the model's trainable parameters. Starting from an initial state \(s_0\), the policy \(\pi_{\theta}\) selects actions \(a_t\) at each timestep \(t\), determining whether the active agent visits a task point or performs a recharging operation based on the current environment state \(s_t\). This iterative decision-making continues until the terminal state \(s_T\) is reached. The output of the trained policy forms a cooperative route \(\mathcal{T}\), comprising a sequence of visited task points and coordinated recharging events between UAVs and UGVs. The routing policy is represented as a joint probability distribution:
\begin{equation}
\mathbb{P}(\mathcal{T} ; \theta) = \prod_{t=0}^{T-1} \pi_\theta(a_t | s_t)
\end{equation}
Here, \(T\) is the total number of timesteps until mission termination. 

\subsection{Encoder-Decoder Transformer Architecture}
The routing policy \(\pi_{\theta}\) leverages an encoder-decoder transformer architecture \cite{vaswani2017attention}, recognized for its robust performance in sequential data processing tasks such as natural language processing, computer vision, and vehicle routing problems \cite{xin2020step, yu2019multimodal, sun2019bert4rec}. This architecture is adapted to process the cooperative routing problem, as illustrated in Fig. \ref{architecture}. The encoder generates high-dimensional embeddings of task points, capturing essential spatial and contextual relationships. The decoder subsequently selects the active agent and determines its actions based on contextual information derived from the current state. This transformer-based approach efficiently models complex interactions between agents and task points, enabling the learning of an optimal routing policy for the UAV-UGV system.

\subsubsection{Encoder}
The encoder utilizes a multi-head attention (MHA) mechanism to transform raw features of the problem instance into high-dimensional representations. The input to the encoder is a 3D vector representation of task points, denoted as \( X = \{o_i = (x_i, y_i, b_i) \,|\, \forall m_i \in \mathcal{M}\} \), where \((x_i, y_i)\) are the normalized coordinates of a task point, and \(b_i\) is a binary indicator denoting whether \(m_i\) is a ground point eligible for UAV-UGV rendezvous ($b_i = 1$ if $m_i$ is ground point and $b_i = 0$ otherwise). Each input vector is initially projected into an embedding space using a linear transformation, $h^0_i = W^0 o_i + b^0$.
Here, \(W^0\) and \(b^0\) are trainable parameters, and the embedding dimension is set to \(d_h = 128\). The initial embeddings \(h^0_i\) are processed through \(L = 3\) attention layers to capture complex task point relationships, yielding the final embeddings \(h^L_i\).

Each attention layer \(l \in \{1, 2, \dots, L\}\) comprises an MHA module, residual connections, a feed-forward (FF) network with ReLU activation, and batch normalization (BN). Within the MHA module, the \textit{Query}, \textit{Key}, and \textit{Value} vectors are computed from the node embeddings of the previous layer. The dimensions of the \textit{Query}/\textit{Key} and \textit{Value} vectors are defined as \(d_q = d_k = d_v = \frac{d_h}{M}\), where \(M = 8\) is the number of attention heads. For each attention head \(j \in \{1, 2, ..., M\}\), attention scores \(Z^l_j\) are computed using the following equations:
\begin{gather}
q^l_{i,j} = h^{l-1}_i W^l_{q,j}, \quad k^l_{i,j} = h^{l-1}_i W^l_{k,j}, \quad v^l_{i,j} = h^{l-1}_i W^l_{v,j} \\
Z^l_j = \text{softmax} \left( \frac{q^l_{i,j} {k^l_{i,j}}^{T}}{\sqrt{d_k}} \right) v^l_{i,j} \\
\text{MHA}(h^{l-1}_i) = \text{Cat}(Z^l_1, Z^l_2, \dots, Z^l_M)
\end{gather}

Here, \(q^l_{i,j}\), \(k^l_{i,j}\), and \(v^l_{i,j}\) correspond to the \textit{Query}, \textit{Key}, and \textit{Value} for attention head \(j\), respectively. The trainable parameter matrices \(W^l_{q,j} \in \mathbb{R}^{d_h \times d_q}\), \(W^l_{k,j} \in \mathbb{R}^{d_h \times d_k}\), and \(W^l_{v,j} \in \mathbb{R}^{d_h \times d_v}\) govern these computations. The outputs of all attention heads are concatenated to produce the final output of the MHA module, \(\text{MHA}(h^{l-1}_i)\). The MHA output is passed through a feed-forward (FF) network with a ReLU activation function, followed by residual skip connections and batch normalization (BN) layers. These operations are summarized as:
\begin{gather}
\hat{h}^l_i = \text{BN}(h^{l-1}_i + \text{MHA}(h^{l-1}_i)) \\
h^l_i = \text{BN}(\hat{h}^l_i + \text{FF}(\text{ReLU}(\hat{h}^l_i)))
\end{gather}

After processing through all attention layers, the final node embedding \(h^L_i\) is obtained, which serves as input to the decoder for subsequent processing.

\subsubsection{Decoder}
\label{decoder_sub}

At each decision-making step \(t\), the decoder executes two primary tasks: it first identifies the active agent using an agent selection strategy, and subsequently determines the task point to visit or perform recharging action using an action selection strategy.

\textbf{Agent Selection Strategy:}
\label{agent_sel_strategy}

\begin{figure}[htp]
\centering
\includegraphics[scale=0.5
]{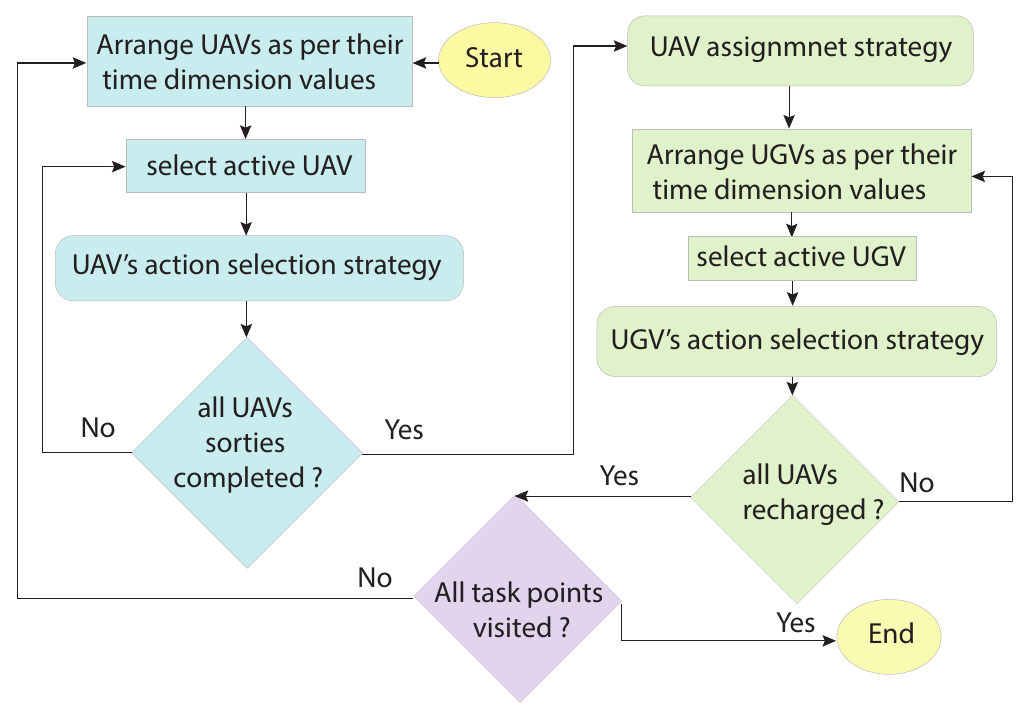}
\caption{Agent selection strategy during the mission progress }
\label{agent_selection}
\end{figure}

The agent selection strategy adopts a sortie-wise agent switching mechanism to improve coordination. This approach allows each agent to complete an entire sortie (defined as the sequence of actions from one recharge event to the next), before switching to another agent. It refines the one-agent-per-decoding-step strategy proposed previously in \cite{fan2022deep}, transitioning from alternating agents at every decision point to a more structured sortie-based routine.

At the beginning of the mission, UAVs are given priority as active agents. The UAV with the smallest time dimension value is selected first. This UAV completes its sortie by visiting multiple task points and concluding at a refueling node. Subsequently, the next UAV, determined by the smallest remaining time dimension value, is selected to begin its sortie. This process continues until all UAVs have completed their sorties. 

Once all UAV sorties are completed, the UGVs are prioritized as active agents. The UGV with the smallest time dimension value is selected first to ensure efficient scheduling. In scenarios with multiple UGVs, UAV assignment becomes critical to determine which UAVs each UGV will service. A greedy UAV assignment policy is employed, where UAVs are assigned to UGVs based on proximity. Specifically, UAVs whose landing locations are closest to a UGV's current position are allocated to that UGV. Each UGV then executes its sortie by visiting its assigned UAVs' refueling points and recharging the respective UAVs. The sequence in which refueling points are visited within a UGV's sortie is determined by the action selection strategy. After all UGV sorties are completed, the UAVs are selected again for their subsequent round of sorties, maintaining the same time-based prioritization. This alternating process where UAVs complete their sorties sequentially, followed by UGVs continues iteratively until all task points are visited, completing the mission as illustrated in Fig. \ref{agent_selection}. 

We can understand this agent selection mechanism with an example of a 3 UAVs–2 UGVs system. Suppose UAV0 has a time dimension value of \(t = 105 \, \text{min}\), UAV1 has \(t = 100 \, \text{min}\), UAV2 has \(t = 110 \, \text{min}\), UGV0 has \(t = 50 \, \text{min}\), and UGV1 has \(t = 60 \, \text{min}\). The agent selection sequence would prioritize UAV1 first, followed by UAV0, and then UAV2. Once all UAV sorties are completed, UGV0 is selected next, followed by UGV1. If the greedy UAV assignment policy assigns UAV0 and UAV2 to UGV0 (based on proximity) and UAV1 to UGV1, UGV0 will visit and recharge UAV0 and UAV2 (with the order determined by the action selection strategy), while UGV1 will recharge UAV1. This strategy ensures efficient synchronization between UAVs and UGVs, reducing idle time and optimizing overall mission performance.

\textbf{Action Selection Strategy:}

Once the active agent is selected, the decoder determines the probability of selecting each available node as an action based on the encoder’s node embedding \( h^L_i \) and a \textbf{context} vector \( h^c_t \), which encapsulates the current scenario state. Separate strategies are employed for UAVs and UGVs to construct the context vector based on their unique operational characteristics.

For a UAV, the \textbf{context} vector \( h^c_t \) is constructed using the UAV’s current position embedding \( h^L_{j,t} \), where \( j \in p_t \) denotes the current position of the agent, the mission status \( d^i_t \), the UAV’s fuel level \( f_t \), and the encoder’s node embeddings \( h^L_i \). Initially, the node embedding \( h^L_i \) is concatenated with the mission status \( d^i_t \), and the resulting vector is linearly projected. A mean graph embedding \( \bar{h}_t \) is then computed by averaging these projections across all nodes. Additionally, the UAV’s current position embedding \( h^L_{j,t} \) is concatenated with its fuel level \( f_t \), and this concatenated vector is linearly projected. The final context vector \( h^c_t \) is obtained by combining the mean graph embedding and the projected UAV state, as follows:
\begin{gather}
\scalebox{1.0}{$\bar{h}_t = \frac{1}{n}\sum_{i=0}^n(\text{Cat}(h^L_i, d^i_t)W_g)$} \\
\scalebox{1.0}{$h^c_t = \bar{h}_t + \text{Cat}(h^L_{j,t}, f_t)W_c$}
\end{gather}

Here, \( W_g \) and \( W_c \) are tunable weight matrices for linear projection, and \(\text{Cat}\) denotes the concatenation operation. This formulation ensures that the context vector captures both the global state of the mission and the local status of the active agent, thereby facilitating accurate decision-making.

Once the context vector \( h^c_t \) is constructed, the decoder uses a multi-head attention (MHA) mechanism to calculate the glimpse vector \( h^g_t \). The context vector is treated as the \textit{Query}, while the encoder’s node embeddings serve as the \textit{Key} and \textit{Value}, as expressed below:
\begin{gather}
h^g_t = \text{MHA}(h^c_t, \ h^L_i W^g_k, \ h^L_i W^g_v)
\end{gather}

Here, \( W^g_k \) and \( W^g_v \) are tunable weight matrices for projecting the node embeddings. The resulting glimpse vector \( h^g_t \) is subsequently used as the \textit{Query} \( q_t \), while the encoder’s node embeddings \( h^L_i \) serve as the \textit{Key} \( k_t \) in a single-head attention mechanism. This mechanism computes the compatibility score \( h_t \) for each node, which determines the likelihood of selecting the corresponding node as action. Importantly, infeasible actions are masked based on logical constraints derived from the current scenario state. The masking criteria include: 
1) previously visited task points (for \textit{visiting} actions), 
2) task points unreachable with the current fuel level of the UAV and 
3) task points that would prevent the UAV from reaching a refuel stop after visiting. The compatibility scores \( h_t \) are computed as follows:
\begin{gather}
\scalebox{1.0}{$q_t = h^g_t W_q, \quad k_t = h^L_i W_k$} \\
\scalebox{1.0}{$h_t = \begin{cases} 
C_p \cdot \tanh \left(\frac{q_t k_t^T}{\sqrt{d_q}}\right) & \text{if feasible} \\ 
-\infty & \text{otherwise.}
\end{cases}$}
\end{gather}

Here, \( W_q \) and \( W_k \) are trainable weight matrices, and \( C_p = 10 \) is a clipping parameter to encourage exploration during training. Finally, the probabilities of selecting actions are derived using the softmax function:
\begin{gather}
\pi_{\theta}(a_t | s_t) = \text{softmax}(h_t)
\end{gather}

In the case of a UGV as the active agent, the decoder constructs the context vector \( h^c_t \) by integrating spatial and temporal information about its assigned UAVs. The UAV assignments are determined using a greedy UAV assignment scheme, which assigns UAVs to UGVs based on proximity. The decoder utilizes the landing locations of the assigned UAVs, represented by their corresponding node embeddings from the encoder \( h^L_i \), as well as the UAVs' landing times and the UGV's current time. 

The process begins by passing the UAVs' landing time information through a Leaky ReLU activation function to generate the landing time embedding \( h^{lt}_i \). Concurrently, the current time of the active UGV is linearly projected and processed through a ReLU activation layer to produce the UGV's current time embedding \( h^{t}_{ugv} \). These two embeddings are combined element-wise to form a unified time vector \( h^{time} \). Subsequently, the landing location embeddings \( h^{loc}_i \) of the assigned UAVs are concatenated with \( h^{time} \), followed by a linear projection and ReLU activation to generate the final context vector \( h^c_t \). The computations are as follows:

\begin{gather}
h^{lt}_i = \text{LeakyReLU}(\text{LandingTime}_i W_{lt}), \ \forall i \in \text{assigned UAVs}, \\
h^{t}_{ugv} = \text{ReLU}(\text{CurrentTime}_{ugv} W_t), \\
h^{time} = h^{lt}_i + h^{t}_{ugv}, \\
h^c_t = \text{ReLU}(\text{Cat}(h^{loc}_i, h^{time}) W_c),
\end{gather}

where \( W_{lt} \), \( W_t \), and \( W_c \) are trainable weight matrices, and \(\text{Cat}\) represents the concatenation operation.

Once the context vector \( h^c_t \) is constructed, it is cross-multiplied with the encoder node embedding \( h^L_i \) to compute the vector \( h_t \). The resulting vector \( h_t \) undergoes mean pooling across its second dimension to calculate the probability distribution over the action space. Similar to the UAV case, infeasible actions are masked to ensure valid decision-making. Specifically, only mission nodes where the active UGV's assigned UAVs have landed are considered feasible action nodes. The probability distribution is computed over this constrained action space, and an action is selected to determine the task point where the UGV will go to recharge the UAV at the specified node.

Following the above process, the decoder selects the active agent and sequentially determines actions to visit task points or perform recharging until the mission is completed. We employ two decoding strategies: a greedy strategy that consistently selects the action with the highest probability at each decision-making step, and a sampling strategy that chooses actions based on their probabilities. During training, we adopt the sampling strategy to encourage better exploration. In the evaluation phase, we assess and compare the effectiveness of both strategies.

\subsection{Training method} 
The training process, as illustrated in Algorithm \ref{algo1}, employs the REINFORCE policy gradient method \cite{williams1992simple} for optimizing the routing policy. This method uses two neural networks: the policy network \(\pi_{\theta}\), responsible for calculating the action probability distribution and sampling actions, and the baseline network \(\pi_{\phi}\), which has the same architecture as the policy network but selects actions greedily, choosing the action with the highest probability.

During each training iteration, the algorithm computes routes and their associated rewards for a batch of problem instances. For the same instances, expected baseline rewards are determined using the greedy rollout generated by the baseline network (lines 4-13). The parameters of the policy network are then updated using the policy gradient method (lines 14-16). To ensure consistency and improve baseline accuracy, the parameters of the baseline network \(\phi\) are updated with those of the policy network after each epoch if the baseline network performs worse than the policy network based on a paired t-test (lines 17-19).

By iteratively updating both the policy and baseline networks, the training algorithm achieves an optimal routing policy over time. This iterative process allows the policy network to continually enhance its decision-making ability through reinforcement learning, while the baseline network serves as a comparative benchmark to refine the updates and guide the learning process effectively.

\vspace{-2mm}
\begin{algorithm}
\footnotesize
\DontPrintSemicolon
\caption{Policy network training using REINFORCE algorithm} \label{algo1}  
    
\KwIn{Policy network $\pi_\theta$, Baseline network $\pi_{\phi}$, epochs $E$, Number of batches $N$, batch size $B$, episode length $T$, UAV-UGV system $\mathcal{A}, \mathcal{G}  $}
\KwOut{Trained policy network $\pi_{\theta^{'}}$}

\For{epoch in $1 \dots E$}{
    Sample $N$ batches from dataset\;
    
    \For{iteration in $1 \dots N$}{
        \For{instance $b$ in $1 \dots B$}{
            Initialize $s_{0,b}$ at $t = 0$ \;
            \While{$t < T$}{
                Get action $a_{t,b} \sim \pi_\theta(a_{t,b} | s_{t,b}) $\;
                Obtain reward $r_{t,b}$ and $s_{t+1, b}$ \;
                $t = t + 1$\;
            }
            
            $ \text{Calculate return} \ \mathcal{R}_b$\;
            Baseline return $\mathcal{R}^{\phi}_b$ from greedy rollout with $\pi_{\phi}$\;
        }
        
        Compute gradient: \vspace{-1mm}\[\nabla_\theta J\leftarrow \frac{1}{B} \sum_{b=1}^B(\mathcal{R}_b-\mathcal{R}_b^{\phi}) \nabla_\theta \log \pi_\theta\left(s_{T, b} \mid s_{0, b}\right)\] \vspace{-4mm} \; 
        
        Update $\theta \leftarrow \theta - \alpha \nabla_\theta J$\;
    }
    
    \If{OneSidedPairedTTest$(\pi_{\theta}, \pi_{\phi}) < 0.05$}{
        $\phi \leftarrow \theta$
    }
}
\end{algorithm}

\section{Results}

To validate the effectiveness of our proposed Deep Reinforcement Learning (DRL) framework for solving the energy-constrained UAV-UGV cooperative routing problem, we conduct extensive computational experiments. These evaluations compare the framework against baseline methodologies, assess its robustness through generalization studies, and demonstrate its practical applicability through a case study involving dynamic task variations and dynamic changes in UAV-UGV team configurations.

\subsection{Dataset Details}

The UAV-UGV cooperative routing problem is simulated over a \(20 \, \text{km} \times 20 \, \text{km}\) operational area with varying UAV-UGV team configurations. UAVs operate at a constant speed of \(v_a = 10 \, \text{m/s}\), while UGVs traverse the road network at \(v_g = 4.5 \, \text{m/s}\). Each UAV is equipped with a fuel capacity of \(F_a = 287.7 \, \text{kJ}\), and its fuel consumption follows a profile modeled after \cite{hurwitz2021mobile}:
\[
\mathcal{P}^a = 0.0461(v_a)^3 - 0.5834(v_a)^2 - 1.8761v_a + 229.6.
\]
This consumption model corresponds to a maximum flight endurance of approximately 25 minutes when flying at \(v_a = 10 \, \text{m/s}\). The UAV-UGV team begins each mission from a common depot. UAVs are tasked with visiting task points located both outside the road network (\(\mathcal{M}_a\)) and on the road network (\(\mathcal{M}_g\)), while UGVs, restricted to the road network, provide logistical support by recharging UAVs at ground points. This ensures continuous mission execution. The task points \(\mathcal{M}_a\) are uniformly sampled within a 7-kilometer radius around road points \(\mathcal{M}_g\). Although the UGV road network \(G\) remains fixed, the specific road points are randomly selected for each instance to introduce variability across simulations. The framework is evaluated on two problem sizes. The smaller problem instance, denoted as U15G5, involves 15 UAV task points and 5 ground points. The larger instance, U45G15, includes 45 UAV task points and 15 ground points. For both problem sizes, separate models are trained for four UAV-UGV team configurations: 1 UAV and 1 UGV, 2 UAVs and 1 UGV, 2 UAVs and 2 UGVs, and 4 UAVs and 2 UGVs. 

Each model is trained on a total of 5,120,000 instances, processed in batches of 256 across 200 batches for 100 epochs. The training data is generated dynamically, ensuring a diverse and robust set of scenarios for each epoch. The Adam optimizer is used with an initial learning rate of \(10^{-4}\), which decays at the end of each epoch at a rate of \(\alpha = 0.995\). The decayed learning rate is computed as:
\[
lr_{\text{decayed}} = lr \times \alpha^{n_{\text{epoch}}}
\]
where \(lr\) is the current learning rate, \(lr_{\text{decayed}}\) is the decayed learning rate, and \(n_{\text{epoch}}\) is the epoch number. Training is performed on an NVIDIA RTX 4090 Ti GPU, with hyperparameters kept consistent across all problem sizes and team configurations. 


\begin{figure}[htp]
\centering
\includegraphics[scale=0.3
]{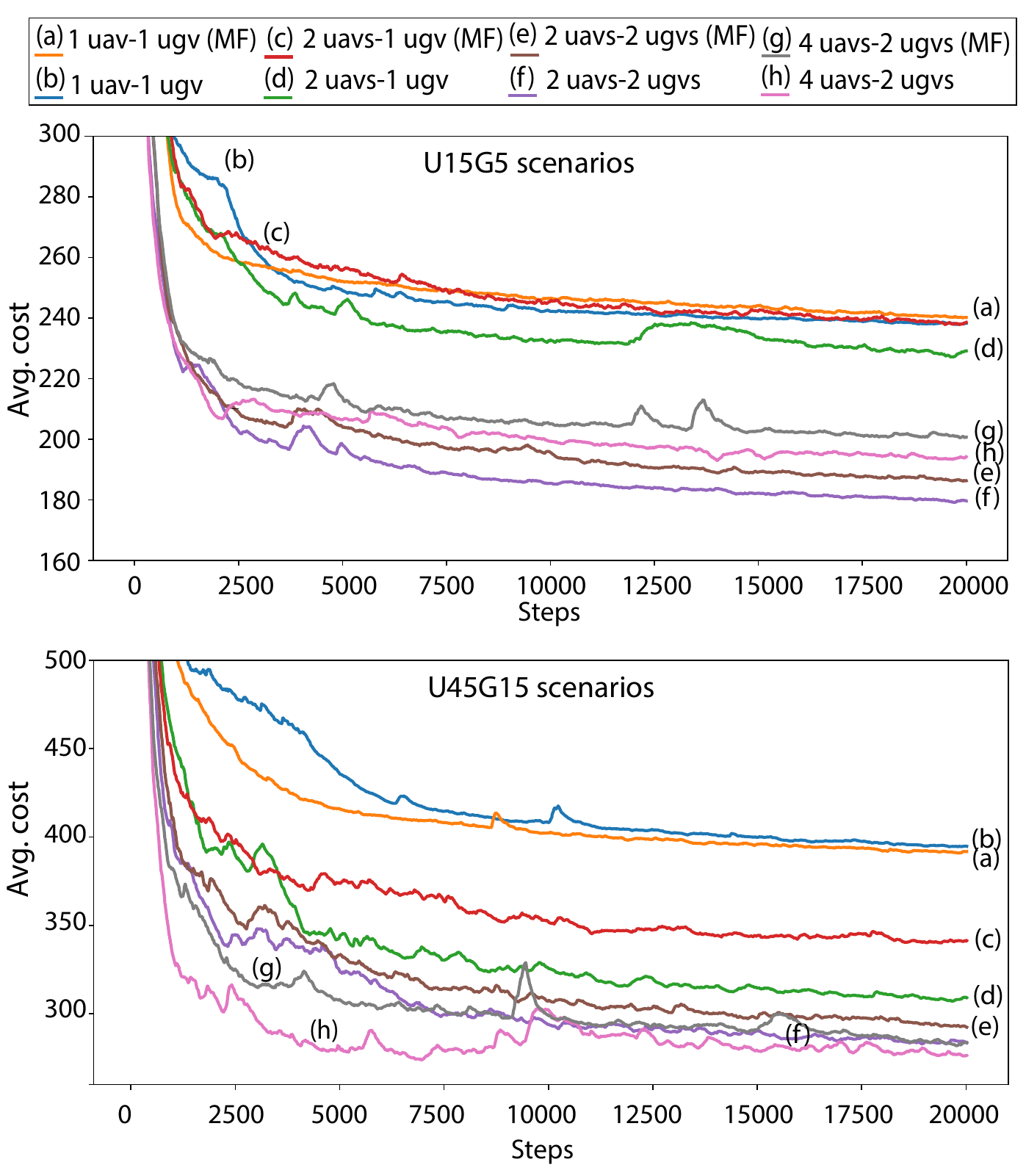}
\caption{Training curves across two problem sizes and different UAV-UGV team compositions. DRL(MF) is a baseline DRL model with the agent selection strategy of Fan et al. \cite{fan2022deep}}
\label{training_curve}
\end{figure}

\subsection{Comparative analysis}

Given the complexity and specificity of the UAV-UGV cooperative routing problem, standard benchmarks are unavailable, and deriving exact solutions becomes increasingly impractical as the number of task points and agents grows, making the problem intractable. Over the years, multi-level or multi-echelon optimization strategies have been widely employed for cooperative routing involving heterogeneous agents. These strategies decompose the problem into smaller subproblems, which are then solved optimally using standard solvers. Following this approach, we adopt a bi-level optimization strategy as the heuristic baseline for addressing our multi-agent cooperative task, as detailed in Section \ref{sec3}. Within this framework, we incorporate three metaheuristic methods: 1) \textbf{Guided Local Search (GLS)}, 2) \textbf{Tabu Search (TS)}, and 3) \textbf{Simulated Annealing (SA)} to solve the E-VRPTW subproblems. These methods serve as heuristic baselines to compare against the performance of our proposed deep reinforcement learning (DRL) framework.

To further analyze the impact of our proposed one-agent-per-sortie agent selection strategy within the decoder of the DRL framework, we include an alternative agent selection strategy, one-agent-per-decoding-step, as proposed in \cite{fan2022deep} as another baseline. In Fan et al.'s approach \cite{fan2022deep}, the decoder selects the agent at each decision step \(t\) based on the index \(t \, \% \, |A|\), where \(A\) is the set of UAVs and \(|A|\) represents the number of agents. For a fair comparison, we retain the encoder and the overall decoder architecture of our DRL framework while substituting the agent selection routine with Fan et al.'s method. The Fig. \ref{training_curve} shows the training curves across two problem sizes with the four team compositions with these two agent selection strategies.

\begin{table*}[h]
\caption{Comparison evaluation of the DRL policy across problem sizes (all metric values represent averages over the test instances)}
\renewcommand{\arraystretch}{1.05}
\tiny
\resizebox{\textwidth}{!}{%
\begin{tabular}{cc|cccccccccccc}
\hline
 & \multirow{3}{*}{\textbf{Model}} & \multicolumn{12}{c}{\textbf{Team configuration}} \\
 &  & \multicolumn{3}{c|}{\textbf{1 UAV-1UGV}} & \multicolumn{3}{c|}{\textbf{2 UAVs-1UGV}} & \multicolumn{3}{c|}{\textbf{2 UAVs-2UGVs}} & \multicolumn{3}{c}{\textbf{4UAVs-2UGVs}} \\ \cline{3-14} 
 &  & \begin{tabular}[c]{@{}c@{}}Obj. \\ (min.)\end{tabular} & \begin{tabular}[c]{@{}c@{}}Gap\\ (\%)\end{tabular} & \multicolumn{1}{c|}{\begin{tabular}[c]{@{}c@{}}Time \\ (sec)\end{tabular}} & \begin{tabular}[c]{@{}c@{}}Obj. \\ (min.)\end{tabular} & \begin{tabular}[c]{@{}c@{}}Gap\\ (\%)\end{tabular} & \multicolumn{1}{c|}{\begin{tabular}[c]{@{}c@{}}Time \\ (sec)\end{tabular}} & \begin{tabular}[c]{@{}c@{}}Obj. \\ (min.)\end{tabular} & \begin{tabular}[c]{@{}c@{}}Gap\\ (\%)\end{tabular} & \multicolumn{1}{c|}{\begin{tabular}[c]{@{}c@{}}Time \\ (sec)\end{tabular}} & \begin{tabular}[c]{@{}c@{}}Obj. \\ (min.)\end{tabular} & \begin{tabular}[c]{@{}c@{}}Gap\\ (\%)\end{tabular} & \begin{tabular}[c]{@{}c@{}}Time \\ (sec)\end{tabular} \\ \hline

\multirow{9}{*}{\rotatebox[origin=c]{90}{{\textbf{Problem size: U15G5}}}} & \textbf{DRL (greedy)} & 238.2 ± 41.2 & 9.0 & \multicolumn{1}{c|}{0.5} & 229.7 ± 59.5 & 15.0 & \multicolumn{1}{c|}{0.5} & 180.5 ± 48.5 & 16.9 & \multicolumn{1}{c|}{0.5} & 188.9 ± 38.7 & 18.1 & 0.4 \\
 & \textbf{DRL MF et al. ( greedy)} & 239.9 ± 42.0 & 9.7 & \multicolumn{1}{c|}{0.4} & 242.7 ± 137.6 & 21.5 & \multicolumn{1}{c|}{0.6} & 184.1 ± 30.8 & 19.3 & \multicolumn{1}{c|}{0.5} & 199.7 ± 37.3 & 24.9 & 0.5 \\
 & \textbf{DRL (1024)} & 220.4 ± 37.9 & 0.8 & \multicolumn{1}{c|}{1.0} & 203.1 ± 34.7 & 1.7 & \multicolumn{1}{c|}{0.9} & 157.1 ± 24.3 & 1.8 & \multicolumn{1}{c|}{1.3} & 163.5 ± 31.3 & 2.2 & 0.9 \\
 & \textbf{DRL MF et al. ( 1024)} & 221.4 ± 38.3 & 1.3 & \multicolumn{1}{c|}{1.0} & 208.4 ± 38.4 & 4.4 & \multicolumn{1}{c|}{0.8} & 158.2 ± 22.8 & 2.4 & \multicolumn{1}{c|}{1.0} & 165.6 ± 25.4 & 3.4 & 0.9 \\
 & \textbf{DRL (10240)} & \textbf{218.6 ± 37.6} & 0.0 & \multicolumn{1}{c|}{7.3} & \textbf{199.7 ± 33.9} & 0.0 & \multicolumn{1}{c|}{8.3} & \textbf{154.4 ± 23.5} & 0.0 & \multicolumn{1}{c|}{8.0} & \textbf{159.9 ± 30.8} & 0.0 & 11.4 \\
 & \textbf{DRL MF et al. ( 10240)} & 219.9 ± 37.9 & 0.6 & \multicolumn{1}{c|}{8.9} & 202.7 ± 36.9 & 1.4 & \multicolumn{1}{c|}{8.2} & 155.6 ± 22.7 & 0.9 & \multicolumn{1}{c|}{8.1} & 163.0 ± 24.8 & 1.9 & 9.8 \\
 & \textbf{GLS} & 281.7 ± 43.0 & 28.9 & \multicolumn{1}{c|}{54.5} & 232.8 ± 39.7 & 16.6 & \multicolumn{1}{c|}{63.6} & 188.3 ± 31.3 & 22.0 & \multicolumn{1}{c|}{46.5} & 219.7 ± 49.5 & 37.4 & 53.8 \\
 & \textbf{TS} & 276.6 ± 42.6 & 26.5 & \multicolumn{1}{c|}{49.6} & 230.8 ± 40.9 & 15.6 & \multicolumn{1}{c|}{56.3} & 185.6 ± 32.4 & 20.2 & \multicolumn{1}{c|}{43.5} & 219.0 ± 47.9 & 36.9 & 53.2 \\
 & \textbf{SA} & 283.4 ± 42.2 & 29.6 & \multicolumn{1}{c|}{54.8} & 242.8 ± 38.2 & 21.6 & \multicolumn{1}{c|}{64.9} & 194.1 ± 29.3 & 25.8 & \multicolumn{1}{c|}{48.1} & 220.3 ± 46.4 & 37.8 & 52.8 \\ \hline

 \multirow{9}{*}{\rotatebox[origin=c]{90}{{\textbf{Problem size: U45G15}}}} & \textbf{DRL (greedy)} & \multicolumn{1}{l}{401.0 ± 50.1} & \multicolumn{1}{l}{12.8} & \multicolumn{1}{l|}{0.7} & \multicolumn{1}{l}{305.4 ± 33.2} & \multicolumn{1}{l}{11.5} & \multicolumn{1}{l|}{0.8} & \multicolumn{1}{l}{277.4 ± 34.8} & \multicolumn{1}{l}{19.3} & \multicolumn{1}{l|}{0.8} & \multicolumn{1}{l}{274.8 ± 105.0} & \multicolumn{1}{l}{25.3} & \multicolumn{1}{l}{1.8} \\
 & \textbf{DRL MF et al. (greedy)} & \multicolumn{1}{l}{399.4 ± 46.5} & \multicolumn{1}{l}{12.3} & \multicolumn{1}{l|}{1.2} & \multicolumn{1}{l}{338.8 ± 40.7} & \multicolumn{1}{l}{23.7} & \multicolumn{1}{l|}{0.8} & \multicolumn{1}{l}{284.6 ± 68.0} & \multicolumn{1}{l}{22.4} & \multicolumn{1}{l|}{1.1} & \multicolumn{1}{l}{275.8 ± 58.6} & \multicolumn{1}{l}{25.7} & \multicolumn{1}{l}{2.0} \\
 & \textbf{DRL (1024)} & \multicolumn{1}{l}{360.8 ± 41.2} & \multicolumn{1}{l}{1.5} & \multicolumn{1}{l|}{3.0} & \multicolumn{1}{l}{276.5 ± 24.9} & \multicolumn{1}{l}{0.9} & \multicolumn{1}{l|}{3.7} & \multicolumn{1}{l}{236.6 ± 25.5} & \multicolumn{1}{l}{1.8} & \multicolumn{1}{l|}{4.6} & \multicolumn{1}{l}{224.4 ± 23.7} & \multicolumn{1}{l}{2.3} & \multicolumn{1}{l}{4.3} \\
 & \textbf{DRL MF et al. (1024)} & \multicolumn{1}{l}{360.2 ± 38.7} & \multicolumn{1}{l}{1.3} & \multicolumn{1}{l|}{2.7} & \multicolumn{1}{l}{285.5 ± 27.7} & \multicolumn{1}{l}{4.2} & \multicolumn{1}{l|}{4.9} & \multicolumn{1}{l}{245.7 ± 24.3} & \multicolumn{1}{l}{5.7} & \multicolumn{1}{l|}{3.3} & \multicolumn{1}{l}{227.4 ± 21.7} & \multicolumn{1}{l}{3.6} & \multicolumn{1}{l}{4.0} \\
 & \textbf{DRL (10240)} & \multicolumn{1}{l}{\textbf{355.6 ± 40.7}} & \multicolumn{1}{l}{0.0} & \multicolumn{1}{l|}{23.0} & \multicolumn{1}{l}{\textbf{273.9 ± 24.6}} & \multicolumn{1}{l}{0.0} & \multicolumn{1}{l|}{23.1} & \multicolumn{1}{l}{\textbf{232.5 ± 24.8}} & \multicolumn{1}{l}{0.0} & \multicolumn{1}{l|}{26.0} & \multicolumn{1}{l}{\textbf{219.3 ± 23.2}} & \multicolumn{1}{l}{0.0} & \multicolumn{1}{l}{37.3} \\
 & \textbf{DRL MF et al. (10240)} & \multicolumn{1}{l}{355.8 ± 37.9} & \multicolumn{1}{l}{0.1} & \multicolumn{1}{l|}{24.7} & \multicolumn{1}{l}{281.0 ± 27.6} & \multicolumn{1}{l}{2.6} & \multicolumn{1}{l|}{28.3} & \multicolumn{1}{l}{240.9 ± 24.1} & \multicolumn{1}{l}{3.6} & \multicolumn{1}{l|}{25.5} & \multicolumn{1}{l}{220.6 ± 21.9} & \multicolumn{1}{l}{0.6} & \multicolumn{1}{l}{34.9} \\
 & \textbf{GLS} & \multicolumn{1}{l}{503.6 ± 54.2} & \multicolumn{1}{l}{41.6} & \multicolumn{1}{l|}{94.6} & \multicolumn{1}{l}{348.5 ± 38.9} & \multicolumn{1}{l}{27.2} & \multicolumn{1}{l|}{101.8} & \multicolumn{1}{l}{284.8 ± 30.3} & \multicolumn{1}{l}{22.5} & \multicolumn{1}{l|}{66.3} & \multicolumn{1}{l}{290.6 ± 43.8} & \multicolumn{1}{l}{32.5} & \multicolumn{1}{l}{82.0} \\
 & \textbf{TS} & \multicolumn{1}{l}{480.4 ± 51.6} & \multicolumn{1}{l}{35.1} & \multicolumn{1}{l|}{85.4} & \multicolumn{1}{l}{354.3 ± 38.2} & \multicolumn{1}{l}{29.4} & \multicolumn{1}{l|}{90.1} & \multicolumn{1}{l}{274.9 ± 33.3} & \multicolumn{1}{l}{18.2} & \multicolumn{1}{l|}{56.6} & \multicolumn{1}{l}{290.4 ± 41.4} & \multicolumn{1}{l}{32.4} & \multicolumn{1}{l}{81.8} \\
 & \textbf{SA} & \multicolumn{1}{l}{507.1 ± 57.2} & \multicolumn{1}{l}{42.6} & \multicolumn{1}{l|}{95.1} & \multicolumn{1}{l}{360.7 ± 41.2} & \multicolumn{1}{l}{31.7} & \multicolumn{1}{l|}{104.2} & \multicolumn{1}{l}{292.3 ± 34.4} & \multicolumn{1}{l}{25.7} & \multicolumn{1}{l|}{71.3} & \multicolumn{1}{l}{293.5 ± 41.8} & \multicolumn{1}{l}{33.8} & \multicolumn{1}{l}{82.9} \\ \hline

\end{tabular}%
}
\label{Table1}
\end{table*}

During evaluation, the DRL-based frameworks employ two decoding strategies: a) greedy decoding, where the action with the highest probability is selected at each decision step, and b) sampling decoding, where \(\mathbb{N}\) trajectories are sampled to form a solution pool, and the best solution is selected. We evaluate the framework with two sampling configurations: $\mathbb{N} = 1024 $ 
  (\textbf{DRL(1024)}) and \(\mathbb{N} = 10240\) (\textbf{DRL(10240)}). To ensure fairness, we apply the same hyperparameters and decoding strategies across our DRL framework and the modified DRL model (\textbf{DRL MF et al.}) based on Fan et al.'s \cite{fan2022deep} agent selection strategy. Given the longer mission durations and the computational requirements of the heuristic approaches, we evaluate on 100 test instances and the average objective value across these instances is used as the primary performance metric to evaluate all baseline methods and our proposed framework for the four UAV-UGV team compositions. All computations are implemented in Python on a Linux system.

Table \ref{Table1} shows that the proposed DRL framework consistently outperforms heuristic-based baselines (GLS, TS, and SA), achieving the lowest average mission time across all problem sizes and team configurations. Among the DRL strategies, sampling-based methods (DRL(1024) and DRL(10240)) demonstrate superior scalability and efficiency, with DRL(10240) delivering the best performance in terms of average objective values. For smaller problem sizes (U15G5), DRL(10240) achieves up to a 9\% reduction in mission time compared to DRL(greedy) in the 1 UAV–1 UGV setup. However, this improvement comes at the cost of significantly higher computation time, making DRL(1024) a more practical alternative. DRL(1024) offers a favorable trade-off, achieving average objective values within 2–3\% of DRL(10240) while requiring up to 7–8 times less computation time. For instance, in the U15G5 configuration with 4 UAVs–2 UGVs, DRL(1024) achieves an objective value only 2.2\% higher than DRL(10240), but requires 10 times less computation time.

Adding more agents generally reduces mission time but exhibits diminishing returns due to recharging constraints because the addition of UAVs increases waiting times for recharging, as one UGV can recharge only one UAV at a time. This effect is more pronounced in smaller problem sizes with fewer task points. For larger problem sizes (U45G15), the DRL framework continues to outperform baseline methods, with DRL(10240) consistently achieving the best objective values. In the U45G15 configuration with 4 UAVs-2 UGVs, DRL(10240) achieves an average objective value of 219.3 minutes, outperforming GLS, TS and SA by 32.5\%, 32.4\% and 33.8\%, respectively. DRL(1024) again provides an efficient alternative, delivering solutions within 2-3\% of DRL(10240) while requiring significantly less computation time. For example, in the U45G15 configuration with 2 UAVs-2 UGVs, DRL(1024) achieves an objective value of 236.6 minutes, just 1.8\% higher than DRL(10240), with less than a third of the computation time.

The DRL framework with the one-agent-per-decoding-step strategy (DRL MF et al.) demonstrates strong performance but is consistently outperformed by the proposed DRL framework. While DRL MF et al.(10240) achieves comparable results, its objective values are slightly higher. For instance, in the U45G15 configuration with 2 UAVs-2 UGVs, DRL MF et al.(10240) achieves an objective value of 240.9 minutes compared to 232.5 minutes for DRL(10240).

Task completion times are generally higher in larger problem sizes due to the increased number of task points and the complexity of agent coordination. For example, in the U15G5 configuration with 1 UAV-1 UGV, DRL(10240) achieves an objective value of 218.6 minutes, which increases to 355.6 minutes in the U45G15 configuration. Heuristic baselines consistently perform worse, with higher objective values and longer computation times. In the U45G15 configuration with 4 UAVs-2 UGVs, GLS delivers an objective value 32.5\% higher than DRL(10240) while requiring more computation time.

Since Table \ref{Table1} presents only the average and standard deviation of the mission time metric, we introduce an additional metric \textit{win rate} to better capture performance variability. \textit{Win rate} is defined as the percentage of test instances where a method achieves the lowest objective value compared to others. In both the U15G5 and U45G15 scenarios, DRL(10240) attains the highest \textit{win rate} across all methods. Detailed \textit{win rate} results, along with the distribution of objective values across the test instances, are provided in the \hyperref[appendix]{Appendix}.





\begin{table*}[!b]
\caption{Comparison evaluation of the DRL policy across
larger problem sizes (all metric values represent averages over the test instances)}
\renewcommand{\arraystretch}{1.05}
\tiny
\resizebox{\textwidth}{!}{%
\begin{tabular}{cc|llllllllllll}
\hline
 & \multirow{3}{*}{\textbf{Model}} & \multicolumn{12}{c}{\textbf{Team configuration}} \\
 &  & \multicolumn{3}{c|}{\textbf{1 UAV-1UGV}} & \multicolumn{3}{c|}{\textbf{2 UAVs-1UGV}} & \multicolumn{3}{c|}{\textbf{2 UAVs-2UGVs}} & \multicolumn{3}{c}{\textbf{4UAVs-2UGVs}} \\ \cline{3-14} 
 &  & \multicolumn{1}{c}{\begin{tabular}[c]{@{}c@{}}Obj. \\ (min.)\end{tabular}} & \multicolumn{1}{c}{\begin{tabular}[c]{@{}c@{}}Gap\\ (\%)\end{tabular}} & \multicolumn{1}{c|}{\begin{tabular}[c]{@{}c@{}}Time \\ (sec)\end{tabular}} & \multicolumn{1}{c}{\begin{tabular}[c]{@{}c@{}}Obj. \\ (min.)\end{tabular}} & \multicolumn{1}{c}{\begin{tabular}[c]{@{}c@{}}Gap\\ (\%)\end{tabular}} & \multicolumn{1}{c|}{\begin{tabular}[c]{@{}c@{}}Time \\ (sec)\end{tabular}} & \multicolumn{1}{c}{\begin{tabular}[c]{@{}c@{}}Obj. \\ (min.)\end{tabular}} & \multicolumn{1}{c}{\begin{tabular}[c]{@{}c@{}}Gap\\ (\%)\end{tabular}} & \multicolumn{1}{c|}{\begin{tabular}[c]{@{}c@{}}Time \\ (sec)\end{tabular}} & \multicolumn{1}{c}{\begin{tabular}[c]{@{}c@{}}Obj. \\ (min.)\end{tabular}} & \multicolumn{1}{c}{\begin{tabular}[c]{@{}c@{}}Gap\\ (\%)\end{tabular}} & \multicolumn{1}{c}{\begin{tabular}[c]{@{}c@{}}Time \\ (sec)\end{tabular}} \\ \hline
 
\multirow{9}{*}{\rotatebox[origin=c]{90}{{\textbf{Problem size: U60G20}}}} & \textbf{DRL (greedy)} & 591.7 ± 339.3 & 35.3 & \multicolumn{1}{l|}{1.3} & 539.3 ± 260.0 & 65.4 & \multicolumn{1}{l|}{1.4} & 389.8 ± 150.2 & 37.7 & \multicolumn{1}{l|}{1.3} & 358.1 ± 96.0 & 33.6 & 1.3 \\
 & \textbf{DRL MF et al. ( greedy)} & 748.8 ± 495.4 & 71.2 & \multicolumn{1}{l|}{1.4} & 651.5 ± 460.1 & 106.4 & \multicolumn{1}{l|}{1.4} & 410.1 ± 220.5 & 44.9 & \multicolumn{1}{l|}{1.4} & 382.4 ± 117.3 & 42.7 & 1.3 \\
 & \textbf{DRL (1024)} & 455.2 ± 45.8 & 4.1 & \multicolumn{1}{l|}{4.0} & 371.3 ± 52.9 & 4.1 & \multicolumn{1}{l|}{4.0} & 295.4 ± 27.2 & 4.4 & \multicolumn{1}{l|}{4.1} & 280.2 ± 24.2 & 4.6 & 4.1 \\
 & \textbf{DRL MF et al. ( 1024)} & 446.6 ± 48.2 & 2.1 & \multicolumn{1}{l|}{4.4} & 380.4 ± 39.3 & 7.4 & \multicolumn{1}{l|}{4.3} & 289.7 ± 26.2 & 2.4 & \multicolumn{1}{l|}{4.4} & 277.9 ± 26.8 & 3.7 & 4.1 \\
 & \textbf{DRL (10240)} & 445.5 ± 44.1 & 1.9 & \multicolumn{1}{l|}{34.2} & \textbf{360.1 ± 48.3} & 0.0 & \multicolumn{1}{l|}{36.3} & 286.6 ± 26.0 & 1.3 & \multicolumn{1}{l|}{40.2} & 269.1 ± 22.4 & 0.4 & 116.2 \\
 & \textbf{DRL MF et al. ( 10240)} & \textbf{437.3 ± 45.6} & 0.0 & \multicolumn{1}{l|}{37.7} & 367.1 ± 39.6 & 2.5 & \multicolumn{1}{l|}{40.5} & \textbf{283.0 ± 25.0} & 0.0 & \multicolumn{1}{l|}{40.5} & \textbf{268.0 ± 25.2} & 0.0 & 116.0 \\
 & \textbf{GLS} & 592.0 ± 58.6 & 35.4 & \multicolumn{1}{l|}{108.2} & 420.6 ± 52.3 & 22.1 & \multicolumn{1}{l|}{122.9} & 336.5 ± 39.2 & 18.9 & \multicolumn{1}{l|}{79.8} & 357.5 ± 59.5 & 33.4 & 97.1 \\
 & \textbf{TS} & 573.6 ± 61.1 & 31.2 & \multicolumn{1}{l|}{99.4} & 429.9 ± 44.9 & 25.4 & \multicolumn{1}{l|}{106.9} & 331.6 ± 37.5 & 17.2 & \multicolumn{1}{l|}{71.8} & 357.3 ± 60.1 & 33.3 & 97.1 \\
 & \textbf{SA} & 602.8 ± 57.6 & 37.8 & \multicolumn{1}{l|}{110.0} & 436.3 ± 53.0 & 27.8 & \multicolumn{1}{l|}{127.4} & 344.1 ± 38.5 & 21.6 & \multicolumn{1}{l|}{82.1} & 359.6 ± 61.4 & 34.2 & 97.8 \\ \hline

\multirow{9}{*}{\rotatebox[origin=c]{90}{{\textbf{Problem size: U75G25}}}}& \textbf{DRL (greedy)} & 673.7 ± 304.6 & 35.3 & \multicolumn{1}{l|}{1.5} & 562.3 ± 200.9 & 40.2 & \multicolumn{1}{l|}{4.0} & 443.1 ± 144.9 & 37.2 & \multicolumn{1}{l|}{1.4} & 388.7 ± 86.3 & 32.2 & 1.4 \\
 & \textbf{DRL MF et al. (greedy)} & 816.5 ± 443.1 & 64.0 & \multicolumn{1}{l|}{1.4} & 692.2 ± 434.2 & 72.6 & \multicolumn{1}{l|}{3.8} & 425.5 ± 168.5 & 31.7 & \multicolumn{1}{l|}{1.4} & 420.9 ± 112.4 & 43.1 & 1.4 \\
 & \textbf{DRL (1024)} & 517.8 ± 42.4 & 4.0 & \multicolumn{1}{l|}{5.8} & 415.9 ± 46.5 & 3.7 & \multicolumn{1}{l|}{4.9} & 333.7 ± 25.9 & 3.3 & \multicolumn{1}{l|}{5.0} & 307.1 ± 26.6 & 4.4 & 4.8 \\
 & \textbf{DRL MF et al. (1024)} & 510.3 ± 42.5 & 2.5 & \multicolumn{1}{l|}{5.1} & 425.1 ± 42.1 & 6.0 & \multicolumn{1}{l|}{5.1} & 330.0 ± 24.1 & 2.2 & \multicolumn{1}{l|}{5.0} & 312.0 ± 28.6 & 5.7 & 4.8 \\
 & \textbf{DRL (10240)} & 503.6 ± 42.8 & 1.2 & \multicolumn{1}{l|}{51.4} & \textbf{401.1 ± 43.5} & 0.0 & \multicolumn{1}{l|}{49.1} & 324.2 ± 26.0 & 0.4 & \multicolumn{1}{l|}{124.3} & \textbf{294.1 ± 23.7} & 0.0 & 47.5 \\
 & \textbf{DRL MF et al. (10240)} & \textbf{497.8 ± 39.5} & 0.0 & \multicolumn{1}{l|}{42.9} & 411.9 ± 40.2 & 2.7 & \multicolumn{1}{l|}{44.8} & \textbf{323.0 ± 22.5} & 0.0 & \multicolumn{1}{l|}{123.2} & 296.6 ± 26.7 & 0.9 & 43.9 \\
 & \textbf{GLS} & 678.2 ± 60.6 & 36.2 & \multicolumn{1}{l|}{120.9} & 462.5 ± 45.2 & 15.3 & \multicolumn{1}{l|}{135.4} & 367.2 ± 40.9 & 13.7 & \multicolumn{1}{l|}{84.9} & 389.3 ± 48.7 & 32.4 & 105.3 \\
 & \textbf{TS} & 654.8 ± 62.8 & 31.5 & \multicolumn{1}{l|}{112.2} & 475.0 ± 47.1 & 18.4 & \multicolumn{1}{l|}{119.5} & 363.3 ± 38.6 & 12.5 & \multicolumn{1}{l|}{79.8} & 390.2 ± 50.8 & 32.7 & 105.8 \\
 & \textbf{SA} & 683.5 ± 60.1 & 37.3 & \multicolumn{1}{l|}{122.5} & 474.9 ± 45.9 & 18.4 & \multicolumn{1}{l|}{138.3} & 385.9 ± 39.0 & 19.5 & \multicolumn{1}{l|}{92.3} & 387.8 ± 51.7 & 31.9 & 105.6 \\ \hline
\end{tabular}%
}
\label{Table2}
\end{table*}

\subsection{Generalization analysis}
To evaluate the generalization capability of the proposed DRL framework, we conduct a series of tests on modified problem scenarios, examining its ability to handle variations without retraining. These tests involve the following aspects:
 \textbf{1. Larger Problem Size}: We increase the problem size beyond the original training size and evaluate the performance of the pre-trained model on these larger scenarios to assess its scalability and adaptability. \textbf{2. Varying Team Composition}: On a given problem scenario, we modify the number of UAVs and UGVs in the team composition and implement the pre-trained model to examine its robustness to changes in the agent configuration. \textbf{3. Different Task Point Distributions:} To simulate diverse operational environments, we create problem scenarios with two different distributions of task points and evaluate the model, which is originally trained on uniformly distributed task points. These generalization experiments aim to test the extrapolation capability of the trained models in handling slightly altered problem scenarios. This analysis ensures that the framework can adapt to conditional changes in real-world applications without requiring retraining. The detailed methodology for each generalization experiment is described below:

\subsubsection{Larger problem size}
In the first experiment, we assess the model’s ability to handle larger problem scenarios than its original training size by creating 100 test instances of two larger sizes. Firstly, we increase the number of task points to a) 60 UAV points and 20 ground points (U60G20) and b) 75 UAV points and 25 ground points (U75G25). Secondly, we extend road network for more available points for recharging. The model’s performance is compared against the baselines on these U60G20 and U75G25 test instances, as listed in Table \ref{Table2}.

The table illustrates that the DRL framework, particularly DRL(10240), achieves competitive mission completion times across all team configurations, demonstrating its generalization capability. However, DRL MF et al.(10240) occasionally outperforms DRL(10240), with an maximum optimality gap of less than 2\%. For example, in the U60G20 configuration with 4 UAVs-2 UGVs, DRL MF et al.(10240) achieves the lowest average objective value of 268.0 minutes, slightly outperforming DRL(10240). Despite this, DRL MF et al.(10240) typically requires more computation time. A similar performance trend is observed in U75G25 scenarios, where DRL methods outperform heuristic baselines. For instance, in the U75G25 configuration with 4 UAVs-2 UGVs, DRL(10240) achieves an average objective value of 294.1 minutes, outperforming GLS, TS, and SA, which record 389.3, 390.2, and 387.8 minutes, respectively. The DRL(1024) variant provides near-optimal results with significantly faster computation times, offering an effective balance between solution quality and runtime efficiency. In general, despite being trained on smaller problem sizes, the DRL framework generalizes well to larger scenarios, achieving competitive results without requiring retraining. In contrast, heuristic baselines (GLS, TS, and SA) exhibit significantly higher mission times and computation costs, highlighting their limitations in handling complex scenarios. Increasing the number of agents reduces mission times across all methods, but the improvement diminishes due to task coordination constraints, especially in scenarios with limited UGVs. DRL models achieve higher \textit{win rate} values compared to heuristic methods. Detailed results and distributions of objective values are provided in the \hyperref[appendix]{Appendix}.

\subsubsection{Team configuration variation}

In the second generalization experiment, we evaluate the performance of a model, trained on a specific UAV-UGV team composition (4 UAVs-2 UGVs), when applied to altered team configurations. Specifically, we test the trained model on the U75G25 problem size with two larger team compositions: 5 UAVs-3 UGVs and 6 UAVs-4 UGVs. Table \ref{Table3} presents the results, highlighting trends in solution quality and runtime efficiency.

For the 5 UAVs-3 UGVs configuration, the proposed DRL framework outperforms all baselines, with DRL(10240) achieving the best solution quality. DRL(10240) records an average mission time of 262.9 minutes, significantly lower than heuristic baselines such as GLS, which achieves 345.6 minutes in 87.6 seconds. DRL(1024) provides near-optimal performance with a mission time of 273.2 minutes in just 5.7 seconds, striking an excellent balance between solution quality and computation time. In comparison, the greedy decoding strategy (DRL(greedy)) offers the fastest runtime (2.0 seconds) but sacrifices solution quality, with an optimality gap of 37.8\% compared to DRL(10240). Heuristic methods (GLS, TS, and SA) have higher mission times and computation costs, with optimality gaps of 6–8\% relative to DRL(10240).

For the 6 UAVs-4 UGVs configuration, the DRL framework outperforms heuristics while being significantly faster, up to 1.5–10 times faster than heuristic approaches. DRL(10240) achieves an average mission time of 263.9 minutes in just 47.2 seconds. DRL(1024) continues to demonstrate efficiency, achieving a mission time of 274.3 minutes in 5.5 seconds, balancing computation time and solution quality effectively. Across both configurations, the proposed DRL framework outperforms the DRL MF et al. model in both solution quality and runtime efficiency. For example, in the 5 UAVs-3 UGVs configuration, DRL(10240) achieves a mission time of 262.9 minutes compared to 282.1 minutes by DRL MF et al.(10240), while being faster computationally. These results highlight the robustness of the proposed DRL framework in adapting to altered team compositions without retraining. More details about the distribution of the mission time values are in the \hyperref[appendix]{Appendix}.

\begin{table}[]
\caption{Performance evaluation of the DRL policy across scenarios with changed team compositions (all metric values represent averages over the test instances)}
\label{Table3}
\renewcommand{\arraystretch}{1.25}
\normalsize
\resizebox{\columnwidth}{!}{%
\begin{tabular}{cc|llllll}
\hline
& \multirow{3}{*}{\textbf{Model}} & \multicolumn{6}{c}{\textbf{Team configuration}} \\
 & & \multicolumn{3}{c|}{\textbf{5 UAVs-3 UGVs}} & \multicolumn{3}{c}{\textbf{6 UAVs-4 UGVs}} \\ \cline{3-8} 
&  & \multicolumn{1}{c}{\begin{tabular}[c]{@{}c@{}}Obj. \\ (sec)\end{tabular}} & \multicolumn{1}{c}{\begin{tabular}[c]{@{}c@{}}Gap\\ (\%)\end{tabular}} & \multicolumn{1}{c|}{\begin{tabular}[c]{@{}c@{}}Time \\ (sec)\end{tabular}} & \multicolumn{1}{c}{\begin{tabular}[c]{@{}c@{}}Obj. \\ (sec)\end{tabular}} & \multicolumn{1}{c}{\begin{tabular}[c]{@{}c@{}}Gap\\ (\%)\end{tabular}} & \multicolumn{1}{c}{\begin{tabular}[c]{@{}c@{}}Time \\ (sec)\end{tabular}} \\ \hline
\multirow{9}{*}{\rotatebox[origin=c]{90}{{\textbf{Problem size: U75G25}}}} & \textbf{DRL (greedy)} & 362.2 ± 128.8 & 37.8 & \multicolumn{1}{l|}{2.0} & 353.8 ± 101.0 & 34.1 & 2.0 \\
 & \textbf{DRL MF et al. (greedy)} & 418.9 ± 176.2 & 59.3 & \multicolumn{1}{l|}{2.1} & 417.3 ± 147.3 & 58.1 & 2.0 \\
 & \textbf{DRL (1024)} & 273.2 ± 20.6 & 3.9 & \multicolumn{1}{l|}{5.7} & 274.3 ± 21.7 & 3.9 & 5.5 \\
 & \textbf{DRL MF et al. (1024)} & 282.1 ± 24.8 & 7.3 & \multicolumn{1}{l|}{6.4} & 288.5 ± 27.8 & 9.3 & 5.4 \\
 & \textbf{DRL (10240)} & \textbf{262.9 ± 20.1} & 0.0 & \multicolumn{1}{l|}{51.8} & \textbf{263.9 ± 19.8} & 0.0 & 47.2 \\
 & \textbf{DRL MF et al. (10240)} & 268.1 ± 22.5 & 2.0 & \multicolumn{1}{l|}{60.7} & 273.5 ± 23.7 & 3.7 & 52.5 \\
 & \textbf{GLS} & 345.6 ± 50.9 & 31.5 & \multicolumn{1}{l|}{87.6} & 308.2 ± 47.0 & 16.8 & 74.2 \\
 & \textbf{TS} & 347.3 ± 49.2 & 32.1 & \multicolumn{1}{l|}{88.1} & 313.0 ± 50.7 & 18.6 & 72.9 \\
 & \textbf{SA} & 352.8 ± 50.3 & 34.2 & \multicolumn{1}{l|}{89.3} & 319.2 ± 51.6 & 21.0 & 76.7 \\ \hline
\end{tabular}%
}
\end{table}

\subsubsection{Task points distribution variation}

\begin{figure}[htp]
\centering
\includegraphics[scale=0.4
]{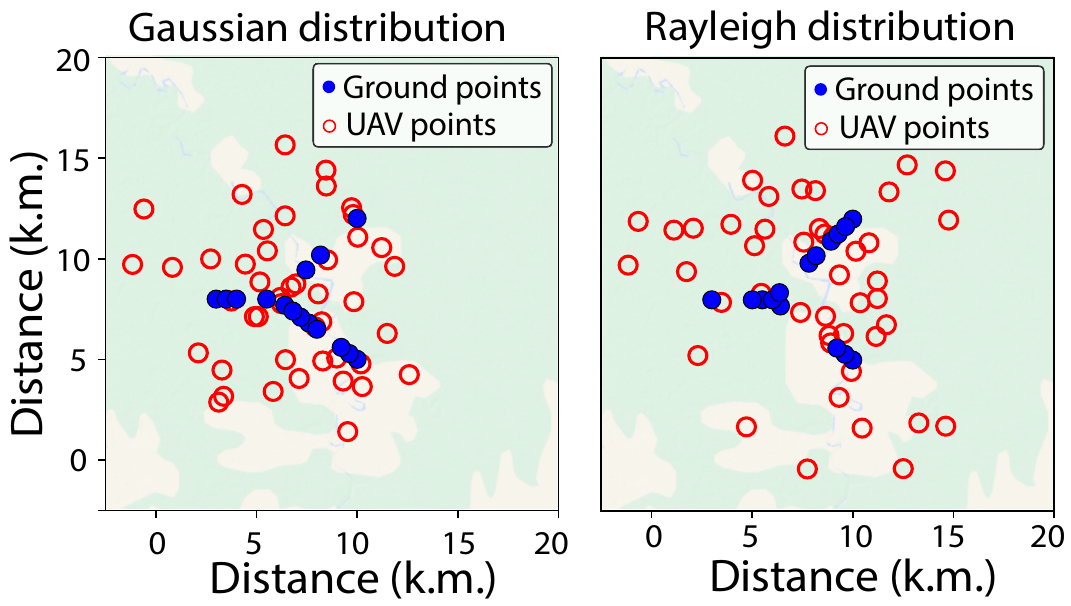}
\caption{Scenario instances from different distributions of UAV points around the road network}
\label{distributions}
\end{figure}

In our third generalization experiment, we assess the adaptability of the proposed DRL policy by testing it on 100 scenarios with two distinct distributions of UAV task points around the road network: (a) Gaussian distribution and (b) Rayleigh distribution. The DRL policy, initially trained on a uniform distribution of task points, is evaluated to determine its capacity to generalize to these alternative distributions. In the Gaussian distribution, UAV task points are concentrated near a central location, with the density decreasing symmetrically as the distance from the center increases. This distribution is particularly suited for surveillance tasks centered around critical areas, such as buildings or urban hubs, where activities are most likely concentrated. In contrast, the Rayleigh distribution features a sparse density of UAV task points near the center, which increases up to a certain radial distance before tapering off. This pattern reflects scenarios where the immediate vicinity of a central location requires minimal surveillance, while the surrounding buffer zone demands higher attention, such as securing areas around critical infrastructure. Fig. \ref{distributions} provides example scenarios for these distributions. These diverse distributions are designed to evaluate the robustness of the DRL policy to varying real-world mission requirements. To test its performance, we use the U45G15 problem size with the four UAV-UGV team configurations and compare the results of the DRL policy against baseline methods. The outcomes of this experiment are summarized in Table \ref{Table4}.

\begin{table*}[h]
\caption{Comparison evaluation of the proposed DRL policy across
variation of UAV task points distribution (all metric values represent averages over the test instances)}
\tiny
\renewcommand{\arraystretch}{1.05}
\label{Table4}
\resizebox{\textwidth}{!}{%
\begin{tabular}{cc|llllllllllll}
\hline
 & \multirow{3}{*}{\textbf{Model}} & \multicolumn{12}{c}{\textbf{Team configuration}} \\
 &  & \multicolumn{3}{c|}{\textbf{1 UAV-1UGV}} & \multicolumn{3}{c|}{\textbf{2 UAVs-1UGV}} & \multicolumn{3}{c|}{\textbf{2 UAVs-2UGVs}} & \multicolumn{3}{c}{\textbf{4UAVs-2UGVs}} \\ \cline{3-14} 
 &  & \multicolumn{1}{c}{\begin{tabular}[c]{@{}c@{}}Obj. \\ (sec)\end{tabular}} & \multicolumn{1}{c}{\begin{tabular}[c]{@{}c@{}}Gap\\ (\%)\end{tabular}} & \multicolumn{1}{c|}{\begin{tabular}[c]{@{}c@{}}Time \\ (sec)\end{tabular}} & \multicolumn{1}{c}{\begin{tabular}[c]{@{}c@{}}Obj. \\ (sec)\end{tabular}} & \multicolumn{1}{c}{\begin{tabular}[c]{@{}c@{}}Gap\\ (\%)\end{tabular}} & \multicolumn{1}{c|}{\begin{tabular}[c]{@{}c@{}}Time \\ (sec)\end{tabular}} & \multicolumn{1}{c}{\begin{tabular}[c]{@{}c@{}}Obj. \\ (sec)\end{tabular}} & \multicolumn{1}{c}{\begin{tabular}[c]{@{}c@{}}Gap\\ (\%)\end{tabular}} & \multicolumn{1}{c|}{\begin{tabular}[c]{@{}c@{}}Time \\ (sec)\end{tabular}} & \multicolumn{1}{c}{\begin{tabular}[c]{@{}c@{}}Obj. \\ (sec)\end{tabular}} & \multicolumn{1}{c}{\begin{tabular}[c]{@{}c@{}}Gap\\ (\%)\end{tabular}} & \multicolumn{1}{c}{\begin{tabular}[c]{@{}c@{}}Time \\ (sec)\end{tabular}} \\ \hline

 \multirow{9}{*}{\rotatebox[origin=c]{90}{{\textbf{Gaussian distribution}}}}
 & \textbf{DRL (greedy)} & 300.7 ± 31.3 & 12.3 & \multicolumn{1}{l|}{0.7} & 252.4 ± 22.0 & 14.8 & \multicolumn{1}{l|}{0.8} & 213.8 ± 25.3 & 19.4 & \multicolumn{1}{l|}{0.8} & 212.9 ± 22.8 & 19.6 & 0.8 \\
 & \textbf{DRL MF et al. ( greedy)} & 301.4 ± 32.7 & 12.6 & \multicolumn{1}{l|}{0.7} & 271.1 ± 30.4 & 23.3 & \multicolumn{1}{l|}{0.8} & 222.8 ± 25.0 & 24.5 & \multicolumn{1}{l|}{0.8} & 225.5 ± 57.2 & 26.6 & 1.2 \\
 & \textbf{DRL (1024)} & 271.7 ± 26.4 & 1.5 & \multicolumn{1}{l|}{2.5} & 224.2 ± 20.7 & 2.0 & \multicolumn{1}{l|}{2.6} & 182.8 ± 17.9 & 2.1 & \multicolumn{1}{l|}{2.4} & 184.4 ± 20.4 & 3.5 & 2.7 \\
 & \textbf{DRL MF et al. ( 1024)} & 273.0 ± 25.9 & 2.0 & \multicolumn{1}{l|}{2.4} & 226.3 ± 18.6 & 2.9 & \multicolumn{1}{l|}{2.5} & 191.9 ± 17.9 & 7.2 & \multicolumn{1}{l|}{2.5} & 184.1 ± 20.1 & 3.4 & 2.5 \\
 & \textbf{DRL (10240)} & \textbf{267.8 ± 27.0} & 0.0 & \multicolumn{1}{l|}{25.0} & \textbf{219.9 ± 21.9} & 0.0 & \multicolumn{1}{l|}{29.3} & \textbf{179.0 ± 17.4} & 0.0 & \multicolumn{1}{l|}{23.0} & 178.4 ± 21.6 & 0.2 & 27.1 \\
 & \textbf{DRL MF et al. ( 10240)} & 269.8 ± 25.1 & 0.8 & \multicolumn{1}{l|}{20.5} & 220.7 ± 18.5 & 0.4 & \multicolumn{1}{l|}{24.1} & 187.9 ± 17.1 & 5.0 & \multicolumn{1}{l|}{25.5} & \textbf{178.1 ± 20.8} & 0.0 & 24.2 \\
 & \textbf{GLS} & 391.4 ± 48.1 & 46.2 & \multicolumn{1}{l|}{73.7} & 268.1 ± 36.1 & 21.9 & \multicolumn{1}{l|}{80.3} & 221.8 ± 29.9 & 23.9 & \multicolumn{1}{l|}{52.0} & 228.2 ± 42.4 & 28.1 & 66.1 \\
 & \textbf{TS} & 376.6 ± 47.8 & 40.6 & \multicolumn{1}{l|}{65.2} & 272.0 ± 38.1 & 23.7 & \multicolumn{1}{l|}{70.0} & 217.5 ± 28.1 & 21.5 & \multicolumn{1}{l|}{49.4} & 227.7 ± 37.9 & 27.8 & 65.4 \\
 & \textbf{SA} & 395.4 ± 48.2 & 47.6 & \multicolumn{1}{l|}{75.0} & 276.1 ± 39.0 & 25.6 & \multicolumn{1}{l|}{82.6} & 225.3 ± 29.9 & 25.9 & \multicolumn{1}{l|}{53.1} & 230.3 ± 37.9 & 29.3 & 65.5 \\ \hline

\multirow{9}{*}{\rotatebox[origin=c]{90}{{\textbf{Rayleigh distribution}}}}
& \textbf{DRL (greedy)} &   328.9 ± 35.5 & 12.8 & \multicolumn{1}{l|}{0.8} & 268.1 ± 67.7 & 13.0 & \multicolumn{1}{l|}{1.2} & 233.2 ± 26.9 & 19.4 & \multicolumn{1}{l|}{0.8} & 231.5 ± 30.2 & 22.0 & 0.8 \\
 & \textbf{DRL MF et al. (greedy)} & 325.1 ± 31.0 & 11.5 & \multicolumn{1}{l|}{0.8} & 289.8 ± 32.6 & 22.2 & \multicolumn{1}{l|}{0.8} & 246.2 ± 68.3 & 26.1 & \multicolumn{1}{l|}{1.2} & 247.3 ± 90.6 & 30.4 & 1.2 \\
 & \textbf{DRL (1024)} & 296.3 ± 28.9 & 1.6 & \multicolumn{1}{l|}{2.5} & 241.3 ± 18.6 & 1.7 & \multicolumn{1}{l|}{2.4} & 199.3 ± 20.5 & 2.0 & \multicolumn{1}{l|}{2.5} & 200.0 ± 16.3 & 5.5 & 3.0 \\
 & \textbf{DRL MF et al. (1024)} & 298.6 ± 28.7 & 2.4 & \multicolumn{1}{l|}{2.5} & 243.7 ± 22.2 & 2.7 & \multicolumn{1}{l|}{2.5} & 208.6 ± 20.5 & 6.8 & \multicolumn{1}{l|}{2.6} & 194.9 ± 17.3 & 2.8 & 2.7 \\
 & \textbf{DRL (10240)} & \textbf{291.7 ± 28.4} & 0.0 & \multicolumn{1}{l|}{23.6} & \textbf{237.2 ± 19.4} & 0.0 & \multicolumn{1}{l|}{22.2} & \textbf{195.3 ± 19.3} & 0.0 & \multicolumn{1}{l|}{21.8} & 194.4 ± 18.1 & 2.5 & 27.7 \\
 & \textbf{DRL MF et al. (10240)} & 295.1 ± 27.7 & 1.2 & \multicolumn{1}{l|}{24.1} & 239.1 ± 21.7 & 0.8 & \multicolumn{1}{l|}{24.9} & 204.6 ± 19.2 & 4.8 & \multicolumn{1}{l|}{24.7} & \textbf{189.7 ± 16.8} & 0.0 & 33.2 \\
 & \textbf{GLS} & 424.0 ± 54.0 & 45.4 & \multicolumn{1}{l|}{81.0} & 292.8 ± 34.7 & 23.4 & \multicolumn{1}{l|}{87.1} & 240.0 ± 34.5 & 22.9 & \multicolumn{1}{l|}{56.8} & 253.7 ± 37.6 & 33.7 & 71.9 \\
 & \textbf{TS} & 408.3 ± 53.2 & 40.0 & \multicolumn{1}{l|}{72.1} & 298.6 ± 36.5 & 25.9 & \multicolumn{1}{l|}{76.6} & 242.4 ± 29.0 & 24.1 & \multicolumn{1}{l|}{53.7} & 252.6 ± 36.6 & 33.2 & 71.4 \\
 & \textbf{SA} & 429.8 ± 57.4 & 47.3 & \multicolumn{1}{l|}{82.2} & 306.4 ± 36.1 & 29.2 & \multicolumn{1}{l|}{90.4} & 244.7 ± 32.4 & 25.3 & \multicolumn{1}{l|}{58.2} & 257.1 ± 36.1 & 35.5 & 72.7 \\ \hline
\end{tabular}%
}
\end{table*}

The results highlight the consistent superiority of the DRL framework, particularly the DRL(10240) policy, across both Gaussian and Rayleigh distributions. DRL(10240) achieves the lowest mission completion times across most team configurations, significantly outperforming heuristic baselines. For example, in the 2 UAVs-2 UGVs configuration under the Gaussian distribution, DRL(10240) records an average objective value of 179.0 minutes, outperforming TS by 21.5\%. The Gaussian distribution generally results in shorter mission times compared to the Rayleigh distribution due to the centralized clustering of task points. In the 4 UAVs-2 UGVs configuration under the Rayleigh distribution, DRL MF et al.(10240) outperforms DRL(10240) with a 2.5\% optimality gap but requires slightly more computation time. Among the heuristic baselines, GLS and TS provide closer performance to the DRL methods but remain significantly slower and less efficient. SA performs the worst, with higher objective values and longer computation times. As observed in previous trends, within the DRL framework, sampling-based methods (DRL(1024) and DRL(10240)) consistently deliver superior solution quality compared to DRL(greedy), though with increased computation times. DRL(greedy), while the fastest in computation, produces suboptimal solutions, making it suitable for time-critical scenarios where computational speed is prioritized over solution quality. The distributions of the objective values are in the \hyperref[appendix]{Appendix}.

\subsection{Case study}
To evaluate the utility of our proposed framework as an online planner, we conduct a case study using a simulated scenario involving 45 UAV task points distributed across a road network. This scenario demonstrates the framework's capability to manage dynamic changes in task environments and team configurations. The study focuses on two key aspects of dynamic route planning: the addition of new task points during the mission and changes in the UAV-UGV team composition. Based on earlier evaluations, DRL(1024) is selected for this case study as it offers the best balance between solution quality and runtime among DRL-based strategies. The replanning process in both scenarios assumes computations are performed during UAV-UGV recharging operations, ensuring efficient planning with minimal mission interruptions.

\subsubsection{Dynamic change in task scenario}
\begin{figure*}[!h]
\centering
\includegraphics[scale=0.35
]{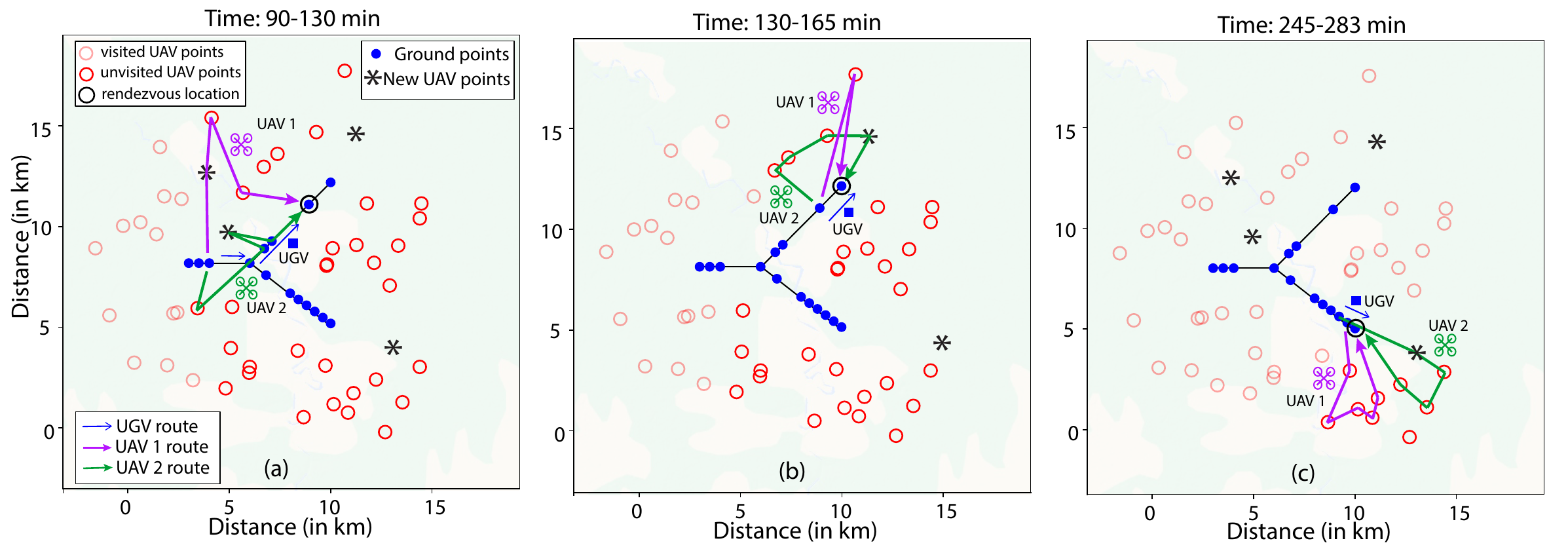}
\caption{Dynamic adaptation of the 2 UAVs–1 UGV system to newly added task points. Less opaque points indicate visited UAV task points, while solid circles represent unvisited points. Newly added points (asterisks) are integrated during UAV-UGV recharging at time 85 minutes. }
\label{dp_1}
\end{figure*}

In the first experiment, new task points are introduced randomly during mission execution. The system triggers a replanning process to integrate these additional points into the existing mission plan. Fig. \ref{dp_1} illustrates the temporal adaptation of UAV route sorties in a 2 UAVs–1 UGV system as new task points are added. The UAVs dynamically adjust their paths, seamlessly incorporating the additional task points into their sorties. This highlights the framework's ability to adapt to evolving task environments without compromising mission efficiency or continuity.

\subsubsection{Dynamic change in team configuration}
\begin{figure*}[!h]
\centering
\includegraphics[scale=0.35
]{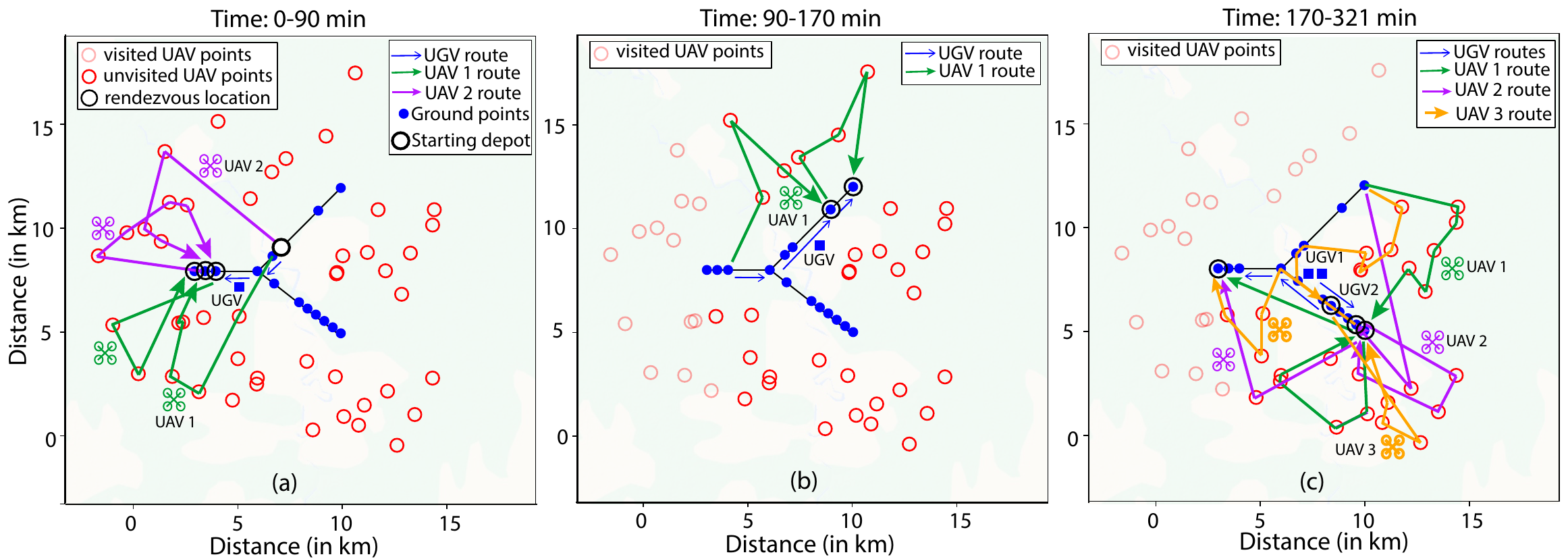}
\caption{Dynamic route planning under varying team configurations. The system adapts to changing team compositions, starting with 2 UAVs–1 UGV, transitioning to reduced and expanded configurations while ensuring all task points are visited efficiently. Less opaque points indicate visited UAV task points, while solid circles represent unvisited points.}
\label{dp_2}
\end{figure*}

In the second experiment, the system adapts to changes in team composition during task execution. The mission begins with a 2 UAVs–1 UGV configuration, which is subsequently modified to both a reduced team of 1 UAV–1 UGV and an expanded team of 4 UAVs–2 UGVs at different stages of the mission. Fig. \ref{dp_2} illustrates the temporal adaptation of route sorties under these varying team configurations. When team size is reduced, the system effectively reassigns task points to the remaining UAVs and UGVs, ensuring mission completion despite fewer resources. Conversely, when additional vehicles are introduced, the framework efficiently scales to accommodate the increased resources, redistributing the task load to optimize coverage and minimize mission time. This adaptability demonstrates the robustness of the framework in dynamic operational scenarios.

\subsection{Discussion \& limitations}

The experimental results demonstrate the scalability, robustness, and generalizability of the proposed DRL framework for UAV-UGV cooperative routing. The framework effectively scales across different team compositions, from a 1 UAV–1 UGV system to a 6 UAV–4 UGV system, maintaining strong performance as problem complexity increases. Across all configurations, it consistently outperforms baseline methods in both solution quality (minimum task completion time) and runtime efficiency. The framework’s generalization capability is validated by extending problem scenarios beyond the original training conditions, including increasing the number of task points, expanding the road network, altering task point distributions, and scaling up team sizes. In all cases, it outperforms baseline approaches, demonstrating its robustness. Furthermore, the case study highlights the framework’s ability to handle online planning, showcasing its adaptability to dynamic changes such as new task points or modified team configurations.

In comparative analyses, the DRL framework with a sampling decoding strategy, particularly the DRL(10240) and DRL(1024) policies, consistently outperforms heuristic baselines such as GLS, TS, and SA in terms of mission completion times across various problem sizes, task distributions, and team configurations. The agent selection strategy in the proposed DRL model proves impactful, as it outperforms the DRL MF et al. model in most instances. While DRL(10240) delivers the best solution quality, DRL(1024) provides a practical trade-off, balancing solution quality with computational efficiency, making it suitable for time-sensitive applications.

The DRL framework surpasses bilevel optimization approaches due to its problem formulation and solution methodology. In bilevel optimization, the problem is addressed hierarchically, where the UGV route is planned first, followed by UAV route sorties. Although this two-stage approach simplifies the problem, it introduces coupling between the stages. Even if both stages produce optimal solutions, the overall mission may not be optimal. For instance, solving E-VRPTW may yield optimal UAV route sorties for individual subproblems but fails to ensure an optimal overall mission route. In contrast, the DRL framework mitigates these limitations by jointly evaluating the entire cooperative route at the end of the mission. Its reward-based feedback mechanism enhances performance by aligning the planning process with global mission objectives rather than isolated subproblem solutions. This unified evaluation ensures a more coherent and effective approach to UAV-UGV collaborative routing, achieving superior mission performance overall.

Despite its promising results, the proposed framework has several limitations. A key limitation is the assumption of a fixed road network for UGV operations. While the generalization tests included extended road layouts, incorporating randomized road networks in future work could improve the model’s robustness. Additionally, the current study is constrained to a maximum of six UAVs and four UGVs due to computational limitations. Scaling to larger team sizes remains a direction for future research. To address this, we propose using curriculum learning by fine-tuning pre-trained models on progressively more complex instances. Another limitation is the use of a greedy strategy for assigning UAVs to UGVs for recharging. Future work will focus on integrating this assignment process directly into the DRL framework to enable more efficient and coordinated decision-making.

\section{Conclusion}

This paper introduces a novel Deep Reinforcement Learning (DRL)-based framework for addressing the energy-constrained multi UAV-UGV cooperative routing problem. By utilizing an encoder-decoder transformer architecture with attention mechanisms, the framework achieves efficient coordination between UAVs and UGVs, ensuring all task points are visited within minimal mission time while adhering to operational constraints. The proposed DRL framework demonstrates significant advantages over heuristic-based bilevel optimization methods, such as Guided Local Search (GLS), Tabu Search (TS), and Simulated Annealing (SA), by delivering superior solution quality and reduced runtime across all problem sizes. Additionally, it exhibits strong generalization capabilities, producing robust solutions in larger scenarios and across diverse task point distributions.
The framework’s adaptability is further highlighted through its performance in dynamic environments, such as scenarios with evolving task points or changing UAV-UGV configurations. The case study underscores the practical applicability of the approach, making it a promising tool for real-world applications, including disaster response and environmental monitoring, where dynamic and efficient multi-agent coordination is critical.

\label{sec:conclusion}

\section*{Acknowledgments}
The authors would like to thank Luca Russo from the University of Illinois Chicago, and James D. Humann and James M. Dotterweich from the DEVCOM Army Research Laboratory for their insightful suggestions. Due to delays in internal approval processes, it was not possible to include them as co-authors at the time of the conference submission. The authors also declare that there are no conflicts of interest among the contributors to this work.



\bibliographystyle{unsrt}
\bibliography{RSS25}

\appendix
\label{appendix}
\counterwithin{figure}{section}
\setcounter{figure}{0}  
\begin{figure*}[h]
    \centering
    \includegraphics[scale=0.4]{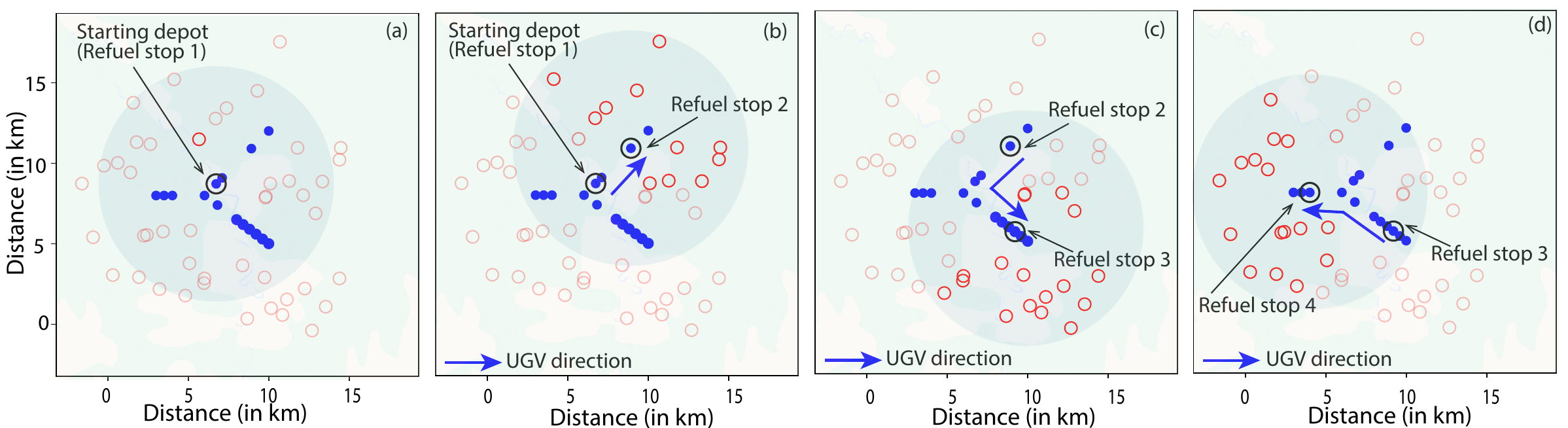}
    \caption{Refuel stops obtained from MSC, with circles indicating the fuel coverage radius of the UAV. (a) Subproblem 1: starting depot (Refuel Stop 1) with allocated UAV points. (b) Subproblem 2: Refuel Stop 1 as the origin and Refuel Stop 2 as the destination with allocated UAV points. (c) Subproblem 3: Refuel Stop 2 as the origin and Refuel Stop 3 as the destination with allocated UAV points. (d) Subproblem 4: Refuel Stop 3 as the origin and Refuel Stop 4 as the destination with allocated UAV points. }
    \label{fig:subproblem_division}
\end{figure*}

\setcounter{figure}{1}  
\begin{figure*}[h]
    \centering
    \includegraphics[scale=0.4]{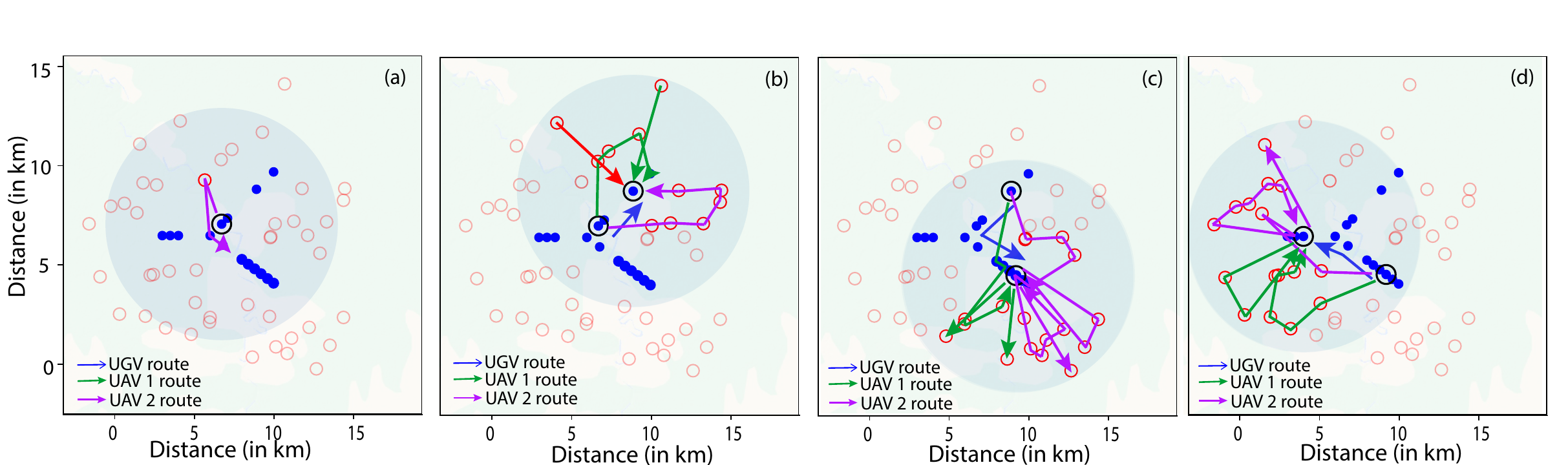}
    \caption{ UAV route sorties as obtained after solving E-VRPTW. (a) UAV route sorties for Subproblem 1. (b) UAV and UGV route sorties for Subproblem 2. (c) UAV and UGV route sorties for Subproblem 3. (d) UAV and UGV route sorties for Subproblem 4. }
    \label{fig:subproblem_division1}
\end{figure*}

\section{Bilevel Optimization Framework}

\subsection{Minimum Set Cover (MSC) Formulation and Subproblem Division for UGV Route Planning}

Maini et al. \cite{maini2015cooperation} demonstrated that in order to establish a
viable cooperative route, it is necessary to ensure that at
least one refueling stop is located within the UAV fuel
coverage radius for each task point. Thus to determine
the minimum number of refueling stops required to cover
the entire mission scenario, we can adopt the minimum set
cover algorithm (MSC).
To determine the minimum number of refueling stops \( \mathcal{R}_e \) required to cover the entire task scenario \( \mathcal{M} \), we employ a linear integer programming formulation solved using a Constraint Programming (CP) method. The problem is modeled with binary decision variables: \( x_j \), which indicates whether a task point \( m_j \) is chosen as a refueling stop, and \( y_{ij} \), which indicates whether a task point \( m_i \) is assigned to a refueling stop \( r_j \). The objective function minimizes the total number of refueling stops:
\[
\text{Minimize:} \quad \sum_{j} x_j
\]
Subject to:
\[
\sum_{j} y_{ij} \geq 1, \quad \forall \, m_i \in \mathcal{M}, \tag{1}
\label{MSC_eq1}
\]
\[
y_{ij} \leq x_j, \quad \forall \, m_i \in \mathcal{M}, \, r_j \in \mathcal{R}_e, \tag{2}
\label{MSC_eq2}
\]
\[
y_{ij} = 0, \quad \text{if} \, d_{ij} > 0.5 F_a, \quad \forall \, m_i \in \mathcal{M}, \, r_j \in \mathcal{R}_e, \tag{3}
\label{MSC_eq3}
\]
\[
x_j, y_{ij} \in \{0, 1\}, \quad \forall \, m_i \in \mathcal{M}, \, r_j \in \mathcal{R}_e. \tag{4}
\label{MSC_eq4}
\]

Here, Equation \ref{MSC_eq1} ensures that each task point \( m_i \) is assigned to at least one refueling stop \( r_j \). Equation \ref{MSC_eq2} ensures that a task point \( m_i \) can only be assigned to a refueling stop \( r_j \) if that stop is selected. Equation \ref{MSC_eq3} excludes task points \( m_i \) from being assigned to refueling stops \( r_j \) that lie beyond the UAV’s coverage radius \( 0.5 F_a \). Equation \ref{MSC_eq4} enforces binary decision variables. We use Google's OR-Tools CP-SAT solver \cite{ORtools} to solve this integer programming problem efficiently. The MSC solution identifies the essential refueling stops \( \mathcal{R}_e \), which are connected by solving traveling salesman problem (TSP) to get the optimal direction of UGVs' traversal.

\setcounter{figure}{2}
\begin{figure*}[!h]
    \centering
    \includegraphics[width=\textwidth]{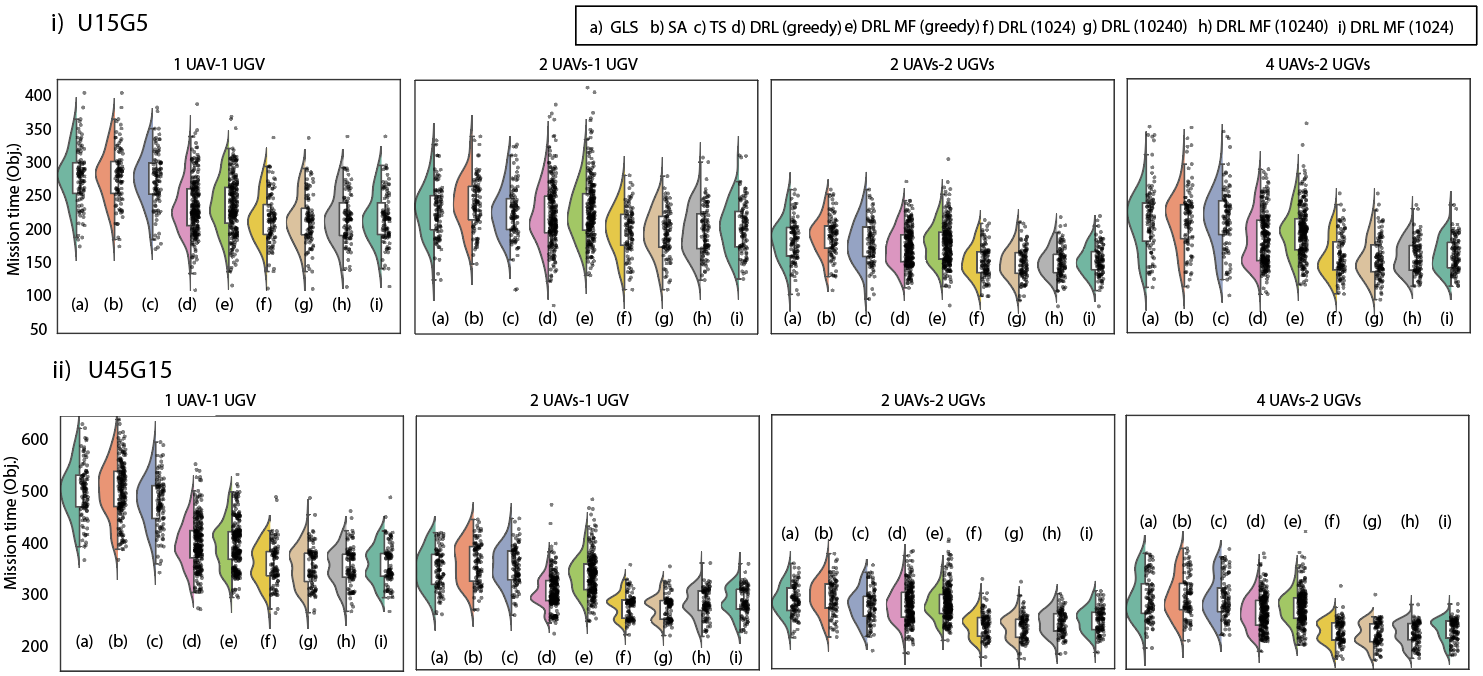}
    \vspace{4mm} 
    \includegraphics[width=\textwidth]{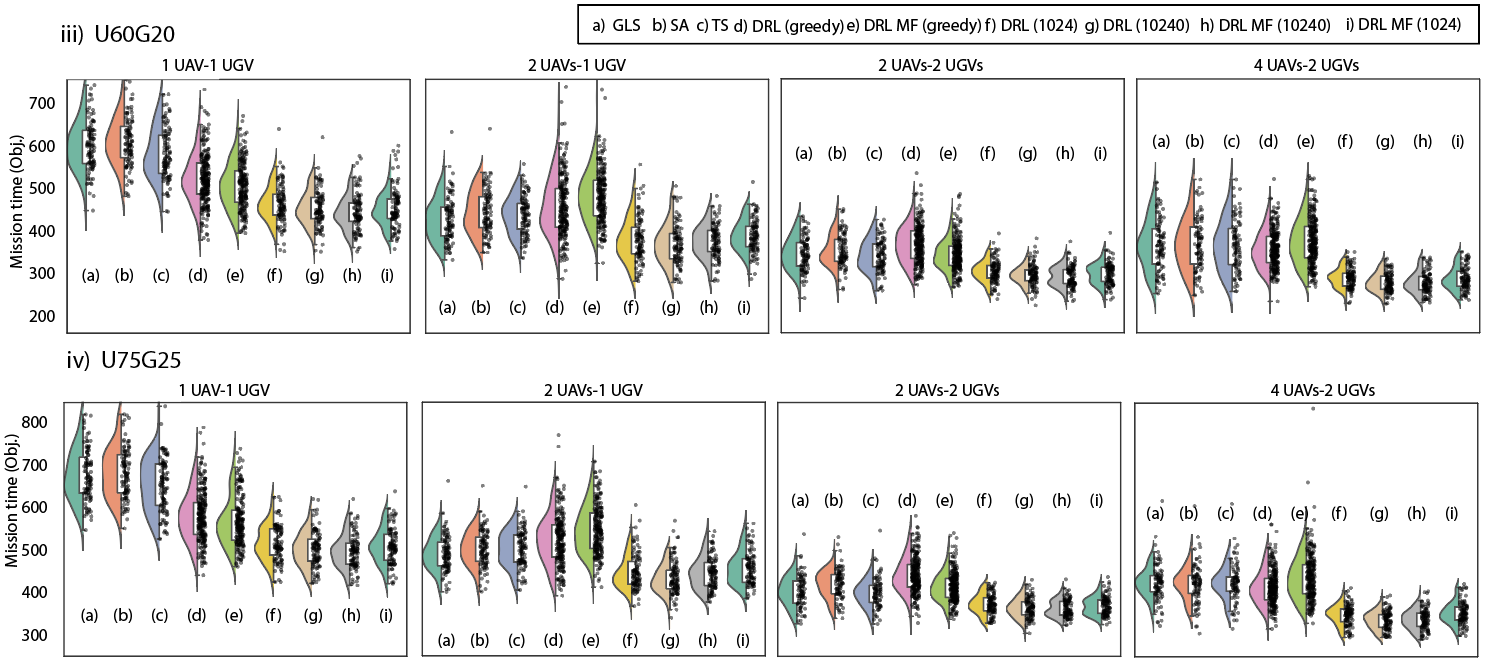}
    \caption{Distribution of mission completion times across test instances for the U15G5, U45G15, U60G20 and U75G25 scenarios. Results are shown for four team configurations and all compared methodologies.}
    \label{fig:a2}
\end{figure*} 

Fig. ~\ref{fig:subproblem_division} illustrates the process of subproblem division and task allocation strategy on the case study scenario. Refueling stops, derived from the MSC solution, are shown in Fig. ~\ref{fig:subproblem_division} and serve as key nodes for structuring UGV routes and partitioning the task space into distinct subproblems. In the first subproblem (Fig.~\ref{fig:subproblem_division}(a)), the starting depot, which also functions as Refuel Stop 1, acts as both the origin and destination node. UAV task points within the coverage radius of this refueling stop are allocated to it. The subsequent subproblems follow a similar pattern. In Subproblem 2 (Fig. ~\ref{fig:subproblem_division}(b)), the starting depot (Refuel Stop 1) becomes the origin node, and Refuel Stop 2 is designated as the destination node. Task points assigned to Refuel Stop 2, based on their proximity, define the task set for this subproblem. This process is repeated for additional subproblems. For instance, in Subproblem 3 (Fig. ~\ref{fig:subproblem_division}(c)), Refuel Stop 2 serves as the origin node, and Refuel Stop 3 is the destination node. Similarly, in Subproblem 4 (Fig. ~\ref{fig:subproblem_division}(d)), Refuel Stop 3 becomes the origin node, and Refuel Stop 4 is the destination. The UGV travels between these refueling stops in each subproblem, establishing time windows at the destination stops for UAVs to land and recharge. 

\subsection{UAV Route Sorties from E-VRPTW}
Considering a 2 UAVs-1 UGV system, The UAV sorties are generated by solving the E-VRPTW model, as described in the problem formulation section. Figure~\ref{fig:subproblem_division1} illustrates the UAV routes obtained after solving the E-VRPTW for each subproblem. In each subproblem, UAVs visit their assigned task points while adhering to fuel constraints. The UAVs recharge at the destination refuel stop before continuing their missions, ensuring energy sustainability throughout the operation. The figure ~\ref{fig:subproblem_division1} highlights the coordinated routing of UAVs and UGVs, where the UGVs establish refueling points while the UAVs efficiently cover task points within the defined subproblem regions.



\setcounter{figure}{3} 
\begin{figure*}[]
    \centering
    \includegraphics[scale=0.41]{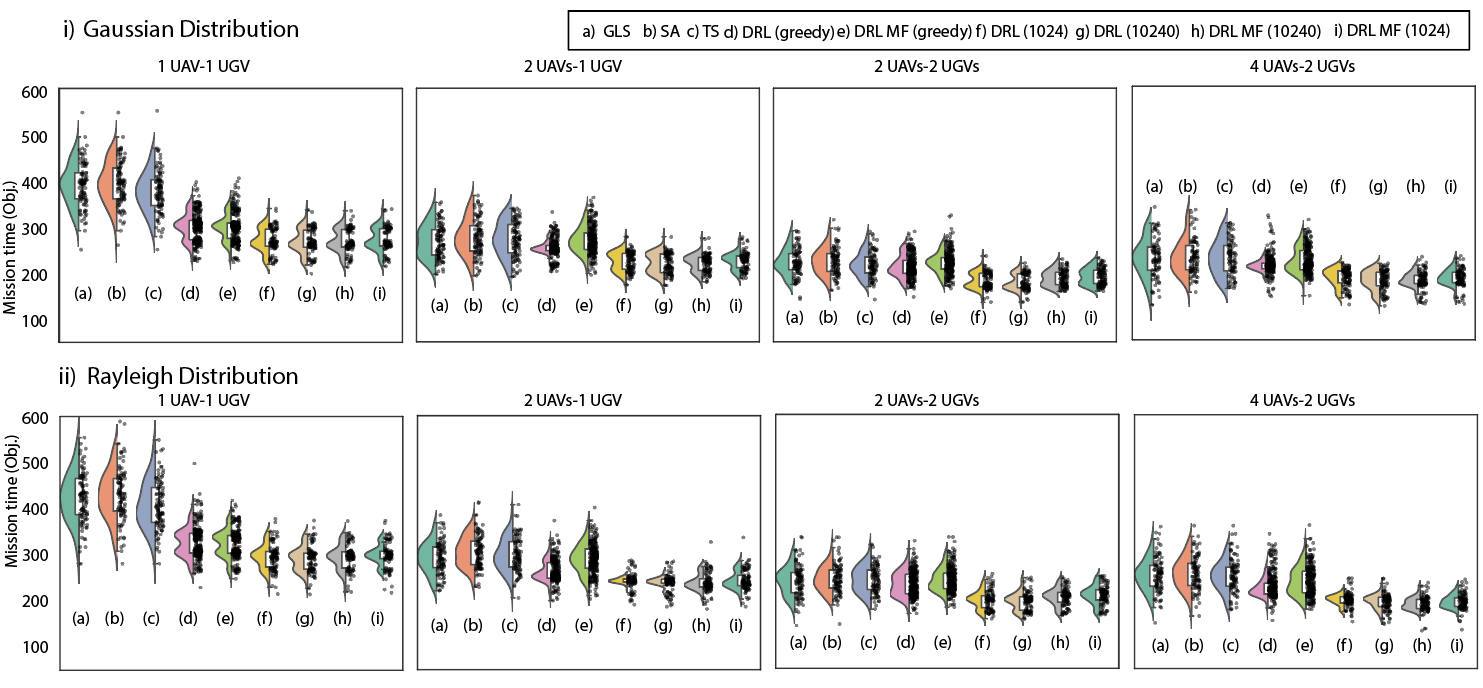}
   \caption{Distribution of mission completion times across test instances for the gaussian and rayleigh distribution scenarios. Results are shown for four team configurations and all compared methodologies.}
    \label{fig:a3}
\end{figure*}

\subsection{Distribution of Objective Values}

To better capture the performance variability of different methods, we compute the \textit{win rate} metric. \textit{Win rate} is defined as the percentage of test instances in which a method achieves the lowest objective value compared to all others. It is formally defined as:
\begin{equation}
\small
\text{\textit{win rate}} = \frac{\text{\# instances where method has the lowest objective}}{\text{total instances}} \times 100
\end{equation}

Overall, DRL-based methods consistently achieve the highest \textit{win rates} across all problem sizes and team configurations, demonstrating strong robustness and generalization capability.

In the U15G5 scenarios, DRL(10240) achieves \textit{win rates} of 68\%, 47\%, 43\%, and 43\% for the 1 UAV–1 UGV, 2 UAVs–1 UGV, 2 UAVs–2 UGVs, and 4 UAVs–2 UGVs configurations, respectively. This is followed by DRL MF et al. (10240), with corresponding \textit{win rates} of 25\%, 42\%, 29\%, and 30\%. Other methods' \textit{win rates} are comparatively less. For the U45G15 scenarios, DRL(10240) obtains \textit{win rates} of 40\%, 50\%, 47\%, and 43\%, while DRL MF et al. (10240) achieves 48\%, 39\%, 19\%, and 42\% across the same configurations.

In the U60G20 scenarios, DRL(10240) records \textit{\textit{win rates}} of 30\%, 56\%, 42\%, and 35\%, whereas {DRL MF et al. (10240)} attains 62\%, 33\%, 26\%, and 33\%, respectively. For the U75G25 scenarios, DRL(10240) achieves \textit{win rates} of 55\%, 55\%, 30\%, and 32\% for the 1 UAV-1 UGV, 2 UAVs–1 UGV, 2 UAVs–2 UGVs, and 4 UAVs–2 UGVs configurations. In comparison, {DRL MF et al. (10240)} achieves 39\%, 32\%, 25\%, and 37\%, respectively.

In the {Gaussian distribution} scenarios, DRL(10240) achieves \textit{win rates} of 49\%, 47\%, 51\%, and 28\% across the 1 UAV–1 UGV, 2 UAVs–1 UGV, 2 UAVs–2 UGVs, and 4 UAVs–2 UGVs configurations, respectively. Meanwhile, DRL MF et al. (10240) achieves 42\%, 46\%, 14\%, and 37\%, respectively. In the Rayleigh distribution scenarios, DRL(10240) obtains \textit{win rates} of 49\%, 41\%, 31\%, and 31\%, while DRL MF et al. (10240) records 41\%, 51\%, 13\%, and 46\%, respectively.

Finally, for larger team configurations with 5 UAVs–3 UGVs and 6 UAVs–4 UGVs, {DRL(10240} achieves \textit{win rates} of 44\% and 42\%, respectively. In comparison, {DRL MF et al. (10240)} achieves 36\% and 20\% in these setups.

\setcounter{figure}{4}  
\begin{figure}[h]
    \centering
    \includegraphics[scale=0.4]{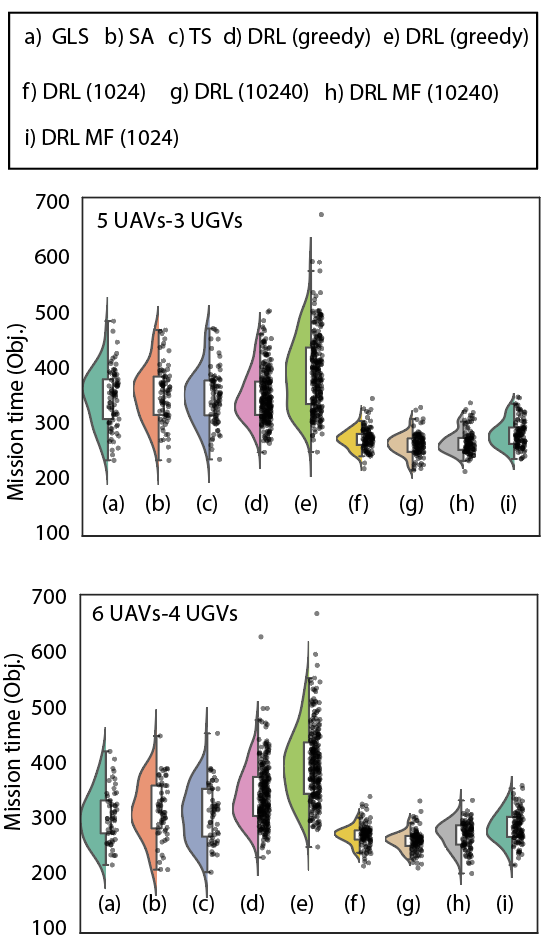}
    \caption{Distribution of mission completion times across test instances for extended team configurations (5 UAVs–3 UGVs and 6 UAVs–4 UGVs) in the U75G25 scenarios.}
    \label{fig:a4}
\end{figure}

Figures~\ref{fig:a2}, \ref{fig:a3}, and \ref{fig:a4} show the distribution of objective values (mission completion times) across all test scenarios. Heuristic methods (GLS, TS, SA) display a wider spread, indicating higher variability, while DRL-based approaches exhibit narrower distributions, reflecting greater consistency. Among DRL methods, the greedy strategy shows more variability than sampling-based methods (DRL(1024), DRL(10240)). Moreover, as the number of agents increases, the distribution width decreases, suggesting improved reliability with larger teams.

\end{document}